\documentclass{article}
\usepackage{ijcai26}

\usepackage{times}
\usepackage{soul}
\usepackage{url}
\usepackage[hidelinks]{hyperref}
\usepackage[utf8]{inputenc}
\usepackage[small]{caption}
\usepackage{graphicx}
\usepackage{amsmath}
\usepackage{amsthm}
\usepackage{booktabs}
\usepackage{algorithm}
\usepackage{algorithmic}
\usepackage[switch]{lineno}

\usepackage{multirow}
\usepackage{makecell}

\usepackage[table]{xcolor}
\usepackage[most]{tcolorbox}
\definecolor{mygray}{gray}{.9}
\definecolor{lightblue}{RGB}{173, 216, 230}
\definecolor{mygreen}{RGB}{179,211,206}
\definecolor{myorange}{RGB}{244,202,199}

\usepackage{amssymb}

\makeatletter
\newcommand\blfootnote[1]{%
  \begingroup
  \renewcommand\@makefnmark{}% 隐藏标记
  \footnotetext{#1}%
  \endgroup
}
\makeatother

\title{Double-Calibration: Towards Reliable LLMs via Calibrating \\ Knowledge and Reasoning Confidence}

\author{
Yuyin Lu$^1$\and
Ziran Liang$^2$\and
Yanghui Rao$^1$\thanks{\textit{Corresponding Author.}}\and
Wenqi Fan$^2$\and
Fu Lee Wang$^3$\And
Qing Li$^2$\\
\affiliations
$^1$School of Computer Science and Engineering, Sun Yat-sen University, Guangzhou, China\\
$^2$Department of Computing, The Hong Kong Polytechnic University, Hong Kong SAR\\
$^3$School of Science and Technology, Hong Kong Metropolitan University, Hong Kong SAR
\emails
luyy37@mail2.sysu.edu.cn,
raoyangh@mail.sysu.edu.cn
}

\begin{document}

\maketitle

\begin{abstract}
    Reliable reasoning in Large Language Models (LLMs) is challenged by their propensity for hallucination. While augmenting LLMs with Knowledge Graphs (KGs) improves factual accuracy, existing KG-augmented methods fail to quantify epistemic uncertainty in both the retrieved evidence and LLMs' reasoning. To bridge this gap, we introduce DoublyCal, a framework built on a novel double‑calibration principle. DoublyCal employs a lightweight proxy model to first generate KG evidence alongside a calibrated evidence confidence. This calibrated supporting evidence then guides a black-box LLM, yielding final predictions that are not only more accurate but also well-calibrated, with confidence scores traceable to the uncertainty of the supporting evidence. Experiments on knowledge-intensive benchmarks show that DoublyCal significantly improves both the accuracy and confidence calibration of black-box LLMs while maintaining low token cost.
\end{abstract}

\section{Introduction}
The reliability of Large Language Models (LLMs) is critically undermined by their tendency to hallucinate~\cite{DBLP:conf/icml/Huang0WWZLGHLZL24}, a problem rooted in both intrinsic \textit{epistemic uncertainty} (knowledge gaps) and extrinsic \textit{aleatoric uncertainty} (data ambiguity)~\cite{DBLP:journals/ml/HullermeierW21}. To mitigate this, Knowledge Graph-augmented Retrieval-Augmented Generation (KG-RAG) has emerged as a leading paradigm~\cite{DBLP:journals/corr/abs-2501-13958}. By augmenting LLM with structured evidence retrieved from external Knowledge Graphs (KGs), KG-RAG is helpful to reduce the model's internal knowledge gaps and improve the factual accuracy of its responses~\cite{DBLP:journals/corr/abs-2506-05690}\blfootnote{This work is to appear in the Proceedings of the 35th International Joint Conference on Artificial Intelligence (IJCAI 2026). A direct link to the conference version will be released at that time.}.

\begin{figure*}[t!]
  \includegraphics[width=\textwidth]{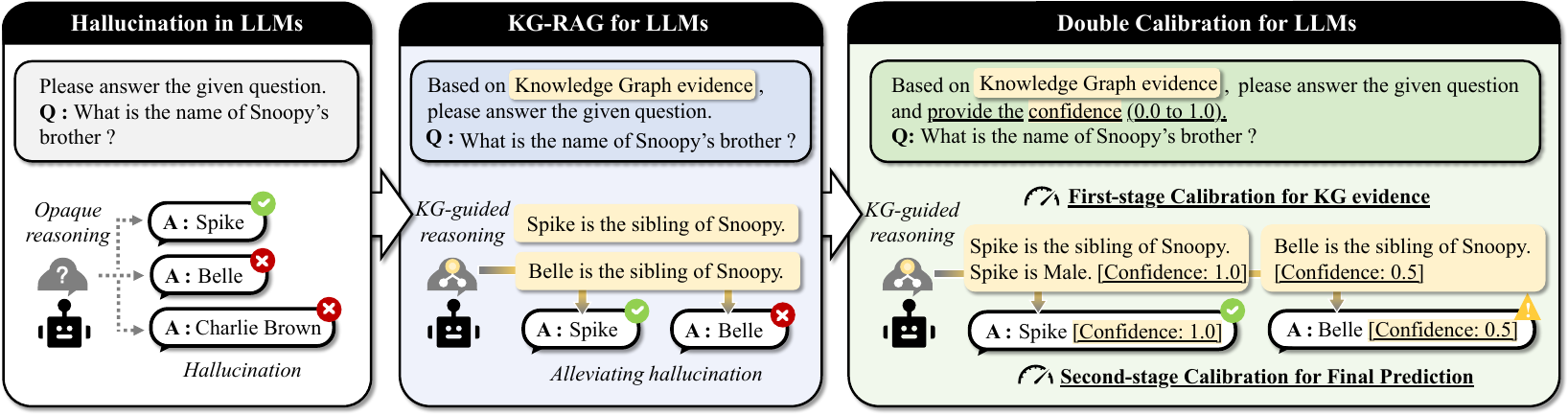}
  \caption{A motivating example of double-calibration against hallucination in KG-augmented LLMs.}
  \label{fig:motivation}
\end{figure*}

However, this KG-augmentation mechanism introduces a new yet critical dependency: \textit{the certainty of the retrieved evidence itself}. Prevailing KG-RAG methods~\cite{DBLP:conf/iclr/LuoLHP24,DBLP:conf/iclr/Li0025,DBLP:journals/corr/abs-2511-10375} often rely on an idealistic assumption that the retrieved evidence is always both sufficient and certain to support correct reasoning for a given query. This assumption is routinely violated in practice due to ambiguous queries, the intrinsic incompleteness of KGs, and imperfections in the retrieval process. Consequently, when provided with partial evidence, LLMs may still produce confidently stated but incorrect predictions~\cite{DBLP:journals/corr/abs-2509-04664}. For example, as illustrated in Figure \ref{fig:motivation}, given the partial evidence ``Belle is a sibling of Snoopy'', an LLM might incorrectly infer ``Belle is Snoopy's brother''. Thus, current KG-RAG lacks the ability to assess and control uncertainty at the very source of the reasoning chain.

Concurrently, research on Uncertainty Quantification (UQ) for LLMs aims to calibrate prediction confidence but focuses predominantly on the final output~\cite{DBLP:conf/acl/XiaXZL25}. For instance, verbalized UQ methods elicit confidence estimates from black-box LLMs, yet these estimates remain opaque and non-traceable~\cite{DBLP:conf/emnlp/TianMZSRYFM23,DBLP:conf/iclr/XiongHLLFHH24}. It is impossible to discern whether the expressed uncertainty stems from flawed evidence, deficiencies in the model's own reasoning, or the intrinsic difficulty of the task. Therefore, existing UQ methods cannot synergize effectively with KG-RAG to provide a stepwise-calibrated view of the complete evidence-to-prediction chain.

In summary, a principled solution for systematically managing the propagation of uncertainty in KG-augmented LLMs is lacking. To bridge this gap, we propose a novel \textbf{double-calibration} paradigm. Its core lies in moving beyond basic evidence retrieval to the construction of \textit{a calibrated reasoning chain}, where confidence is explicitly estimated and made traceable from the retrieved KG evidence to the final LLM prediction. As shown in Figure \ref{fig:motivation}, this enables the LLM to weigh alternative answers (e.g., correctly favoring ``Spike'' over ``Belle'') based on the calibrated confidence of the supporting evidence, rather than making an overconfident guess.

We instantiate this principle in \textbf{DoublyCal}, a framework that implements double-calibrated KG-RAG. DoublyCal grounds the LLM's reasoning on verifiable KG evidence and performs dual calibration: it first calibrates the confidence of the retrieved evidence, then uses this calibrated evidence to guide and further calibrate the final LLM prediction. Specifically, we formalize KG evidence as constrained relational paths extracted from a KG. We then train a lightweight proxy model under Bayesian supervision to generate relevant KG evidence alongside a calibrated confidence score for each query. During inference, the primary LLM is prompted with both the KG evidence and its confidence estimate, leading to more accurate and better-calibrated predictions. Crucially, because the evidence confidence explicitly estimates the expected reasoning uncertainty of the LLM when utilizing the provided evidence, the final confidence becomes traceable to the verifiable KG evidence and its calibrated confidence, rather than remaining an opaque global estimate. Our main contributions are summarized as follows:
\begin{itemize}
    \item We establish the principle of double-calibration for reliable KG-augmented LLMs, which mandates explicit confidence calibration for both the KG evidence and the final LLM predictions.
    \item We propose DoublyCal\footnote{The source code of DoublyCal is available at \url{https://github.com/luyy9apples/DoublyCal}.}, a framework that implements this principle via a Bayesian-calibrated proxy model, providing the primary LLM reasoner with KG evidence accompanied by evidence confidence.
    \item We empirically demonstrate that DoublyCal significantly and consistently improves the accuracy and calibration of diverse black-box LLMs on knowledge-intensive benchmarks in a cost-effective manner.
\end{itemize}

\section{Related Work}
\subsection{Knowledge-Augmented Generation for Reliable LLMs}
Retrieval-Augmented Generation (RAG) reduces the inherent knowledge gaps of LLMs by providing external information, thereby improving the factual accuracy of their responses~\cite{DBLP:journals/corr/abs-2501-13958}. The choice of knowledge source defines a spectrum of RAG variants, ranging from (i) unstructured text in Vanilla RAG~\cite{DBLP:journals/corr/abs-2410-05779,DBLP:journals/corr/abs-2505-07233}, to (ii) textual graphs that model latent connections in GraphRAG~\cite{DBLP:conf/nips/He0SC0LBH24,DBLP:conf/iclr/LiC0L0T0H0L25}, and finally to (iii) formal Knowledge Graphs (KGs) with explicit relations in KG-RAG~\cite{DBLP:conf/iclr/LuoLHP24,DBLP:conf/iclr/Li0025}. By providing precise and structured knowledge, KG-RAG offers a rigorous foundation for complex reasoning and has demonstrated superior performance on knowledge-intensive tasks~\cite{DBLP:journals/corr/abs-2506-05690,DBLP:journals/tkde/PanLWCWW24}.

However, the retrieved knowledge itself may be noisy or insufficient, posing a persistent challenge to the reliability of LLM reasoning. To address this, some research dynamically selects external knowledge when it conflicts with the LLM’s internal parametric knowledge~\cite{DBLP:journals/corr/abs-2511-10375,DBLP:conf/acl/ZhangXXWLWS25}. Unlike prior work that focus on resolving knowledge conflicts, we introduce a double-calibration principle which explicitly quantifies confidence for both the retrieved evidence and the final LLM prediction, thereby systematically identifying epistemic boundaries.

\subsection{Uncertainty Quantification for LLMs}
Uncertainty Quantification (UQ) for LLMs aims to calibrate the confidence of model predictions to identify their epistemic boundaries~\cite{DBLP:conf/icml/Huang0WWZLGHLZL24}. While some studies incorporate uncertainty awareness during training~\cite{stangel2025rewarding}, most practical UQ methods often operate post-hoc and are categorized by model access~\cite{DBLP:conf/acl/XiaXZL25}.

For open‑source LLMs, confidence is typically derived from internal states, such as the feature‑space distribution of hidden embeddings~\cite{DBLP:conf/iclr/0026L0GWTFY24,DBLP:conf/naacl/VazhentsevRLPPBS25} and the predictive entropy of the output distribution~\cite{DBLP:conf/iclr/MalininG21}. For black‑box LLMs, methods rely on API‑based probing. A prevalent strategy generates multiple responses and evaluates their semantic consistency~\cite{DBLP:conf/emnlp/ManakulLG23,DBLP:conf/iclr/KuhnGF23,DBLP:journals/tmlr/LinT024}. A more efficient alternative is verbalized UQ, which directly prompts the LLM to verbalize its own confidence, eliciting introspective uncertainty estimates~\cite{DBLP:conf/emnlp/TianMZSRYFM23,DBLP:conf/iclr/XiongHLLFHH24,DBLP:conf/aistats/TanneruAL24}. Its plug-and-play nature and low cost make verbalized UQ readily integrable with KG-RAG, forming a strong single-calibration baseline that calibrates only the final output.

A fundamental limitation across prior UQ paradigms is their exclusive focus on the final output, deriving confidence solely from the LLM’s internal states or self-assessment. This makes them vulnerable to the model’s overconfidence biases due to the lack of an objective external anchor~\cite{DBLP:conf/iclr/XiongHLLFHH24}. Our double‑calibration principle addresses this by first calibrating externally verifiable KG evidence, thereby establishing a traceable foundation for the entire reasoning chain.

\section{Preliminaries}
\subsection{Uncertainty and Confidence}
\label{sec:def-unc-conf}
We extend the conceptualization of uncertainty and confidence for LLM outputs \cite{DBLP:journals/tmlr/LinT024} to general predictive systems. Given an input $\boldsymbol{x}$, a predictive system $f$ produces a probability distribution over possible outputs $P_f(\boldsymbol{o} \mid \boldsymbol{x})$. The uncertainty of $f$ regarding $\boldsymbol{x}$ is quantified by the dispersion (e.g., entropy) of this distribution. Conversely, the overall confidence of $f$ can be defined inversely to this uncertainty. For a specific output $\boldsymbol{o}_i$, its confidence is directly associated with its assigned probability $P_f(\boldsymbol{o} = \boldsymbol{o}_i \mid \boldsymbol{x})$.

\begin{figure*}[t!]
  \includegraphics[width=\textwidth]{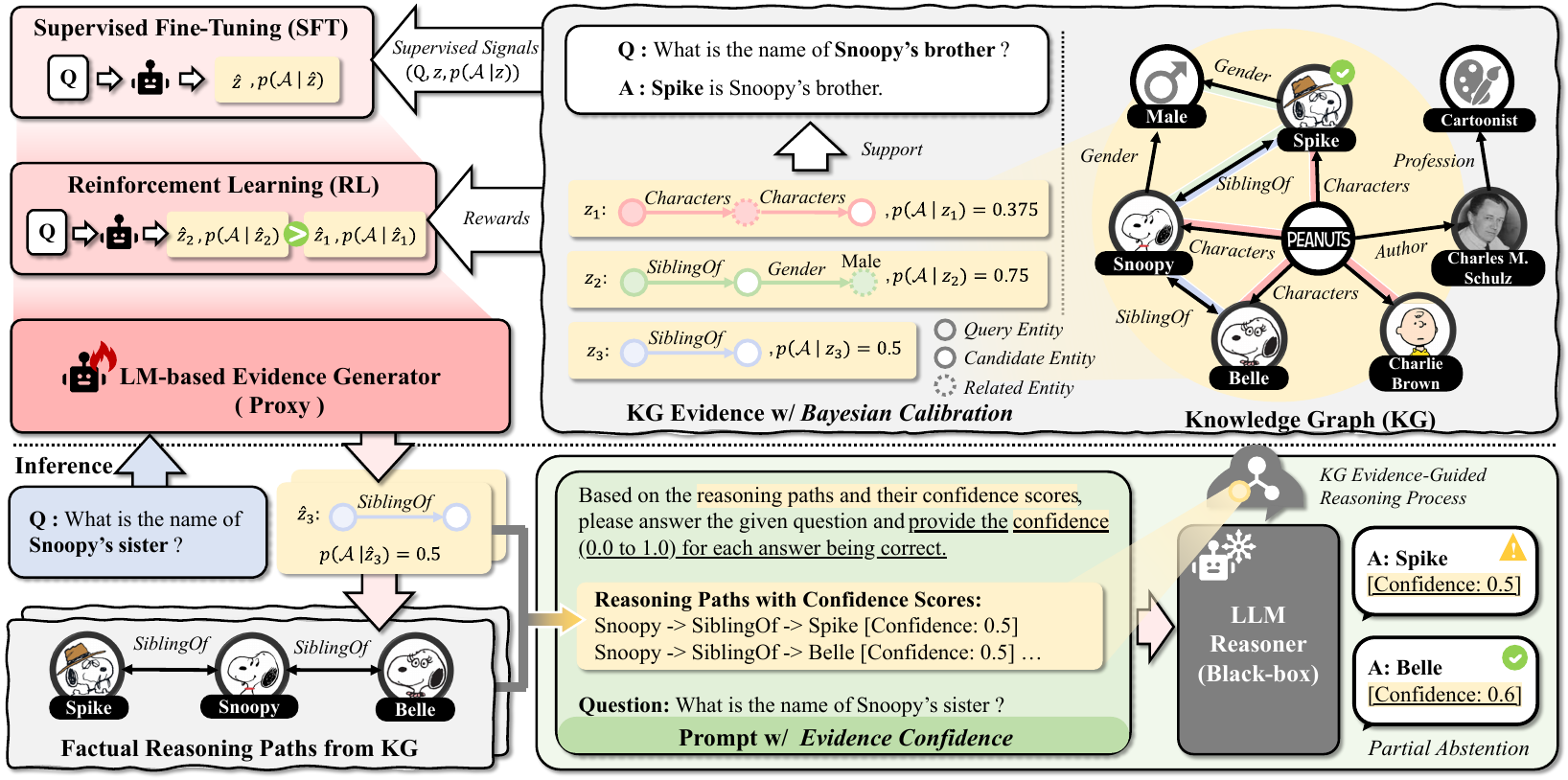}
  \caption{Overview of the DoublyCal framework. (\textbf{Top-right}) Bayesian calibration estimates statistically grounded confidence for KG evidence. (\textbf{Top-left}) A lightweight proxy model is trained to generate calibrated evidence-confidence pairs. (\textbf{Bottom}) During inference, these calibrated pairs guide a black-box LLM reasoner to produce final answers with well-calibrated confidence.}
  \label{fig:DoublyCal}
\end{figure*}

\subsection{Knowledge Graph}
A Knowledge Graph (KG) is a graph-structured database representing factual knowledge as a set of triples~\cite{DBLP:conf/sigmod/BollackerEPST08}. Formally, a KG is denoted as $\mathcal{G}:=(\mathcal{V}, \mathcal{R}, \mathcal{E})$, where $\mathcal{V}$ is a set of entities, $\mathcal{R}$ is a set of relations, and $\mathcal{E}:={(h, r, t)} \subseteq \mathcal{V}\times\mathcal{R}\times\mathcal{V}$ is a set of factual triples. Each triple $(h, r, t)$ represents an atomic fact, stating that relation $r$ holds between head entity $h$ and tail entity $t$. To mitigate the knowledge gaps of LLMs, KG-RAG provides them with structured evidence extracted from KGs~\cite{DBLP:journals/corr/abs-2506-05690}.

\subsection{Knowledge Graph Question Answering} 
Knowledge Graph Question Answering (KGQA) is a canonical knowledge-intensive reasoning task~\cite{DBLP:conf/acl/YihRMCS16}. Given a natural language question $\boldsymbol{Q}$ involving query entities $\mathcal{V}_{\boldsymbol{Q}}$, a reasoning system is expected to retrieve relevant evidence from a KG $\mathcal{G}$ and reason over it to produce the correct answer set $\mathcal{A}$. A standard knowledge-augmented pipeline (e.g., KG-RAG) involves two stages: (i) a retriever $g$ that fetches a set of relevant evidence $\mathcal{Z}_{\boldsymbol{Q}}=g(\boldsymbol{Q}; \mathcal{G})$, and (ii) a reasoner $f$ (e.g., an LLM) that predicts answers $\hat{\mathcal{A}}=f(\boldsymbol{Q};\mathcal{Z}_{\boldsymbol{Q}}, \mathcal{G})$. This decomposition naturally highlights two distinct sources of uncertainty that our framework aims to calibrate: the evidence uncertainty in $\mathcal{Z}_{\boldsymbol{Q}}$, and the reasoning uncertainty in generating $\hat{\mathcal{A}}$ given $\mathcal{Z}_{\boldsymbol{Q}}$.

\section{Methodology}
This section introduces \textbf{DoublyCal}, a framework designed to establish a calibrated reasoning chain by jointly calibrating both verifiable KG evidence and the final LLM predictions.
As illustrated in Figure~\ref{fig:DoublyCal}, the proposed DoublyCal framework operates through three core components. 
Firstly, we formalize KG evidence as constrained relational paths and employ a Bayesian model to estimate a statistically grounded confidence for each evidence (Sec.~\ref{sec:evidence}).
Then, a lightweight proxy model is trained under the supervision of these Bayesian confidence scores to generate KG evidence alongside its calibrated confidence (Sec.~\ref{sec:method-proxy}).
Finally, the calibrated evidence-confidence pair serves as an objective signal integrated into any black-box LLM to mitigate its inherent overconfidence, thereby enhancing both the calibration and traceability of its final predictions (Sec.~\ref{sec:llm-reason}).

\subsection{Bayesian Calibration of KG Evidence}
\label{sec:evidence}
\paragraph{KG Evidence Formulation.} Effective KG evidence must balance \textit{informativeness} for accurate reasoning with \textit{interpretability} for reliable confidence estimation. While relational paths~\cite{DBLP:conf/iclr/LuoLHP24} offer step-by-step interpretability, they may lack sufficient context. In contrast, subgraphs~\cite{DBLP:conf/iclr/Li0025} provide broader context but often introduce redundancy. To resolve this trade-off, we introduce \textit{constrained relational paths} as our primary evidence form. This formulation augments a core relational path with an optional constraint derived from the neighborhood of the candidate answer, thereby enhancing informativeness while preserving interpretability.

Formally, given a KG $\mathcal{G}$ and a question $\boldsymbol{Q}$, a constrained relational path $\mathcal{P}_c$ is defined as the conjunction of a relational path $\mathcal{P}_r$ and an optional constraint $\mathcal{C}$:
\begin{equation}
\mathcal{P}_c := \mathcal{P}_r \left[\,\wedge\,\mathcal{C}\,\right],
\end{equation}
where $\mathcal{P}_r:= \exists v_1, \dots, v_{l-1}. r_1(q, v_1) \wedge r_2(v_1, v_2) \wedge \dots \wedge r_l(v_{l-1}, \hat{a})$ denotes a directed relational path of length $l$ from the query entity $q\in\mathcal{V}_{\boldsymbol{Q}}$ to a candidate answer $\hat{a}$. Here, each $r_i \in \mathcal{R}$ is a relation, and $v_i$ is an existential variable. The optional constraint $\mathcal{C} := r_c(\hat{a}, c)$ represents a one-hop triple from the candidate answer $\hat{a}$ to a constraint entity $c \in \mathcal{V}$, which serves to filter or refine the candidate set.

\paragraph{Example.} \textit{Consider the question ``What is the name of Snoopy's brother?'' with the query entity $q=\texttt{Snoopy}$ and the true answer $a=\texttt{Spike}$. Figure~\ref{fig:DoublyCal} illustrates a relational path evidence $z_3$ and its constrained counterpart $z_2$:}
\begin{align}
z_3 &:= \mathrm{SiblingOf}(q, \hat{a}) \models_{\mathcal{G}} \{\texttt{Spike}, \texttt{Belle}\}, \\
z_2 &:= z_3 \wedge \mathrm{Gender}(\hat{a}, \texttt{Male}) \models_{\mathcal{G}} \{\texttt{Spike}\},
\end{align}
\textit{where $\models_{\mathcal{G}}$ denotes grounding the evidence in $\mathcal{G}$ to obtain candidate answer entities. The auxiliary constraint on the candidate’s gender effectively identifies $\hat{a}=\texttt{Spike}$, yielding a more precise and informative evidence for reasoning.}

\paragraph{Confidence Estimation with Beta-Bernoulli Model.}
To estimate a statistically grounded confidence for a given KG evidence $z_{\boldsymbol{Q}}$, we model it as a predictive system in accordance with the definition in Sec.~\ref{sec:def-unc-conf}. Formally, the system takes as input a candidate answer $\hat{a}$ drawn from the candidate set $[\![z_{\boldsymbol{Q}}]\!]$ obtained by grounding $z_{\boldsymbol{Q}}$ in $\mathcal{G}$ (i.e., $z_{\boldsymbol{Q}} \models_{\mathcal{G}} [\![z_{\boldsymbol{Q}}]\!]$), and produces a binary output indicating whether $\hat{a}$ is correct ($\hat{a} \in \mathcal{A}$). This defines a Bernoulli distribution for the correctness of a uniformly sampled candidate, characterized by the parameter $p \in [0, 1]$, which is precisely the probability that the candidate is correct.

To obtain a robust estimate of $p$ that accounts for KG incompleteness, we impose a conjugate Beta prior $p\sim \mathrm{Beta}(\alpha, \beta)$, with hyperparameters $\alpha, \beta > 0$. Given $\boldsymbol{Q}$ and its answer set $\mathcal{A}$, we use the posterior mean as the calibrated confidence score $p^*$, which has the following closed form:
\begin{equation}
\label{eq:bayesian-map}
p^* = p(\mathcal{A} \mid z_{\boldsymbol{Q}}) = \frac{\alpha + \bigl| [\![z_{\boldsymbol{Q}}]\!] \cap \mathcal{A} \bigr|}{\alpha + \beta + \bigl| [\![z_{\boldsymbol{Q}}]\!] \bigr|},
\end{equation}
where $\bigl| [\![z_{\boldsymbol{Q}}]\!] \cap \mathcal{A} \bigr|$ counts the number of correct candidates, and $\bigl| [\![z_{\boldsymbol{Q}}]\!] \bigr|$ is the total number of grounded candidates. $p^*$ blends the empirical accuracy of the evidence with prior belief, mitigating the impact of sparse KG grounding (See Appendix~\ref{apx:theory-prior} for detailed analysis).

\paragraph{Example (cont.).} \textit{With a weakly informative prior set to $\alpha=\beta=0.5$~\cite{jeffreys1998theory}, the statistical confidence for KG evidence $z_3$ and $z_2$ is estimated as $p(\mathcal{A} \mid z_3) = 0.5$ and $p(\mathcal{A} \mid z_2) = 0.75$.}

\subsection{Proxy for Evidence Generation \& Calibration}
\label{sec:method-proxy}
The Beta-Bernoulli model-based confidence provides a statistically grounded but \textit{retrospective}
measure of evidence quality. To enable \textit{prospective} evidence retrieval and confidence estimation during inference, we introduce a lightweight \textbf{reasoning proxy}. This proxy model is designed to generate high‑quality KG evidence alongside well‑calibrated confidence estimates for any input question, thereby approximating a reliable reasoning path before the LLM's final prediction. The proxy model is implemented by an \textbf{LM-based Evidence Generator} and trained using the Bayesian confidence scores as its supervisory signal.

\paragraph{Supervised Fine-Tuning (SFT) for Evidence and Confidence Generation.}
We formalize the dual task of evidence generation and confidence estimation as a sequence-to-sequence problem and conduct the initial training of the proxy model via SFT. This stage equips the proxy with the fundamental ability to identify relevant KG evidence and estimate confidence by mimicking the Bayesian signal.

The SFT training dataset is constructed from triples $\left( \boldsymbol{Q}, z_{\boldsymbol{Q}}, p(\mathcal{A}\mid z_{\boldsymbol{Q}}) \right)$. Each triple is formatted into a structured sequence using a predefined template: the question serves as the instruction, and the target output is the evidence path enclosed in XML-style tags, with the Bayesian confidence score included as an attribute (e.g., $\texttt{<PATH confidence=}\dots\texttt{>}\dots\texttt{</PATH>}$; {see Appendix~\ref{sec:appendix-prompt-sft} for details}). The proxy model $f_{\theta}$ is trained via standard autoregressive language modeling to generate this target sequence ({see Appendix~\ref{sec:appendix-sft-obj} for the objective}). 

\paragraph{Reinforcement Learning (RL) for Evidence Decision.}
We further refine the proxy by framing evidence generation and calibration as a sequential decision process optimized via RL. This stage transitions the proxy from imitation to strategic decision-making that jointly maximizes both inferential quality and confidence calibration.

For each question $\boldsymbol{Q}$ with a gold evidence set $\mathcal{Z}_{\boldsymbol{Q}}$, we compute a reward for the generated evidence $\hat{z}_{\boldsymbol{Q}}$ and its predicted confidence $\hat{c}$. Firstly, we define a match score $m(\hat{z}_{\boldsymbol{Q}}, z_{\boldsymbol{Q}}) \in [0, 1]$ for each $z_{\boldsymbol{Q}}\in\mathcal{Z}_{\boldsymbol{Q}}$, which combines Jaccard similarity~\cite{jaccard1901etude} with an order-sensitive Levenshtein ratio~\cite{lcvenshtcin1966binary}. The overall reward $R$ is a weighted combination of an inferential quality reward $R_{\text{inf}}$ and a calibration alignment reward $R_{\text{cal}}$:
\begin{align}
& R = \lambda \cdot R_{\text{inf}} + (1 - \lambda) \cdot R_{\text{cal}}, \label{eq:reward}\\
& R_{\text{inf}} = \text{F1}(z_{\boldsymbol{Q}}) \cdot m(\hat{z}_{\boldsymbol{Q}}, z_{\boldsymbol{Q}}), \\
& R_{\text{cal}} = \max \left( 0, 1 - \xi \cdot \left| \hat{c} - c \right| \right), \\
& \text{with } c = p(\mathcal{A}\mid z_{\boldsymbol{Q}}) \cdot m(\hat{z}_{\boldsymbol{Q}}, z_{\boldsymbol{Q}}).
\end{align}

\begin{table*}[t]
\centering
\resizebox{\textwidth}{!}{ 
\begin{tabular}{@{}p{2.6cm}<{\centering}cp{2.2cm} l ccccc l ccccc@{}}
\toprule[1.5pt]
\multirow{2}{*}{\textbf{Reasoning Method}} & \multirow{2}{*}{\textit{KG Evidence}} & \multirow{2}{*}{+ UQ Method} & \multicolumn{5}{c}{\textbf{WebQSP}} & & \multicolumn{5}{c}{\textbf{CWQ}} \\
\cmidrule(lr){4-8} \cmidrule(lr){10-14}
& & & \textbf{Hits} & \textbf{Recall} & \textbf{F1} & \textbf{ECE} $\downarrow$ & \textbf{ACE} $\downarrow$ & & \textbf{Hits} & \textbf{Recall} & \textbf{F1} & \textbf{ECE} $\downarrow$ & \textbf{ACE} $\downarrow$ \\
\midrule
\midrule
\multirow{3}{*}{\makecell{\textbf{LLM Reasoner} \\ (GPT-3.5-turbo)}} & \multirow{3}{*}{\makecell{\textit{No Augmentation}}}
 & + Vanilla & 74.7 & 53.1 & 44.6 & 27.7 & 26.6 & 
 & 47.7 & 40.3 & 29.5 & 38.8 & 38.0 \\
 & & + CoT & 75.4 & 53.9 & \cellcolor{mygray}44.4 & 26.6 & 25.8 & 
 & 48.2 & 41.2 & \cellcolor{mygray}29.3 & 38.4 & 37.6 \\
 & & + Self-Probing & \cellcolor{mygray}74.1 & \cellcolor{mygray}52.8 & 50.2 & \cellcolor{mygray}36.5 & \cellcolor{mygray}36.3  &
 & \cellcolor{mygray}43.6 & \cellcolor{mygray}36.6 & 34.3 & \cellcolor{mygray}48.5 & \cellcolor{mygray}47.9 \\
\midrule
\multirow{3}{*}{\makecell{\textbf{RoG} \\ \cite{DBLP:conf/iclr/LuoLHP24}}}
 & \multirow{3}{*}{\makecell{\textit{Relational Path}}} & + Vanilla & 89.3 & 77.6 & 67.1 & 19.6 & 20.8  &
 & 65.3 & 60.5 & 43.0 & 27.4 & 26.8 \\
 & & + CoT & 89.9 & 78.5 & 68.2 & 18.9 & 17.9 &
 & 65.3 & 60.4 & 43.7 & 27.6 & 27.0 \\
 & & + Self-Probing & 87.5 & 76.6 & 73.5 & 13.9 & 12.8  &
 & 61.9 & 56.9 & 48.7 & 38.0 & 37.4 \\
\midrule
\multirow{3}{*}{\makecell{\textbf{SubgraphRAG} \\ \cite{DBLP:conf/iclr/Li0025}}}
 & \multirow{3}{*}{\makecell{\textit{Subgraph}}} & + Vanilla & 88.8 & 81.3 & \cellcolor{mygreen}77.3$^\ddag$ & 11.1 & 9.7  &
 & 61.5 & 57.4 & \cellcolor{mygreen}52.2$^\ddag$ & 39.9 & 39.7  \\
 & & + CoT & 89.0 & 81.0 & 77.1 & 10.6 & 9.5  &
 & 59.4 & 55.7 & 51.4 & 38.9 & 38.7 \\
 & & + Self-Probing & 89.6 & 80.7 & 74.9 & 12.3 & 12.8  &
 & 59.9 & 56.0 & 50.2 & 39.1 & 38.0 \\
\midrule
\midrule
\multirow{3}{*}{\makecell{\textbf{SFT-DoublyCal} \\ (Ours)}}
 & \multirow{3}{*}{\makecell{\textit{Constrained} \\ \textit{Relational Path}}} & + Vanilla & 90.0 & 81.0 & 72.6 & \cellcolor{myorange}3.1$^\dag$ & \cellcolor{mygreen}{3.8}$^\ddag$  &
 & 68.8 & 64.3 & 48.1 & 17.9 & 17.5 \\
 & & + CoT & \cellcolor{mygreen}90.1$^\ddag$ & 81.3 & 72.1 & \cellcolor{mygreen}3.5$^\ddag$ & \cellcolor{myorange}{3.6}$^\dag$ &
 & 69.0 & 64.7 & 47.7 & \cellcolor{mygreen}17.8$^\ddag$ & 17.4 \\
 & & + Self-Probing & 88.5 & 79.4 & 76.6 & 7.9 & 7.1  &
 & 63.2 & 58.5 & 50.9 & 22.5 & 22.1 \\
\midrule
\multirow{3}{*}{\makecell{\textbf{RL-DoublyCal} \\ (Ours)}}
 & \multirow{3}{*}{\makecell{\textit{Constrained} \\ \textit{Relational Path}}} & + Vanilla & \cellcolor{myorange}91.5$^\dag$ & \cellcolor{mygreen}84.8$^\ddag$ & 76.7 & 4.5 & 5.1  &
 & \cellcolor{mygreen}{70.5$^\ddag$} & \cellcolor{mygreen}66.6$^\ddag$ & 50.1 & 17.9 &\cellcolor{mygreen} 17.3$^\ddag$ \\
 & & + CoT & \cellcolor{myorange}91.5$^\dag$ & \cellcolor{myorange}85.0$^\dag$ & 76.8 & 3.9 & 4.3 &
 & \cellcolor{myorange}71.3$^\dag$ & \cellcolor{myorange}67.5$^\dag$ & 49.8 & \cellcolor{myorange}17.6$^\dag$ & \cellcolor{myorange}17.2$^\dag$ \\
 & & + Self-Probing & 89.9 & 83.0 & \cellcolor{myorange}79.3$^\dag$ & 6.8 & 6.8  &
 & 64.6 & 60.8 & \cellcolor{myorange}53.0$^\dag$ & 23.5 & 23.1 \\
\bottomrule[1.5pt]
\end{tabular}
}
\caption{Main results (\%) of our DoublyCal and the SingleCal baselines on WebQSP and CWQ datasets. Best, second-best, and worst results are highlighted in \colorbox{myorange}{red}$^\dag$, \colorbox{mygreen}{green}$^\ddag$, and \colorbox{mygray}{gray}, respectively.}
\label{tab:main_results}
\end{table*}

Here, $\text{F1}(z_{\boldsymbol{Q}})$ is a precomputed F1 score assessing the reasoning capability of the gold evidence. Intuitively, a lower match score $m(\hat{z}_{\boldsymbol{Q}}, z_{\boldsymbol{Q}})$ reduces both the inferential quality of $\hat{z}_{\boldsymbol{Q}}$ and its target confidence $c$. The weight $\lambda\in (0,1)$ balances the two objectives, and $\xi > 0$ is a tolerance coefficient.
The final reward per generation is the maximum $R$ over $\mathcal{Z}_{\boldsymbol{Q}}$, followed by transformations to ensure a smooth training signal. The policy $\pi_{\theta}$ of the proxy model is optimized to maximize the expected reward under the Group Relative Policy Optimization (GRPO) \cite{DBLP:journals/corr/abs-2402-03300} objective ({see Appendices~\ref{sec:appendix-prompt-rl} and \ref{sec:appendix-rl} for implementation details}).

\subsection{LLM Reasoning with Calibrated Evidence}
\label{sec:llm-reason}
The trained proxy equips any black‑box primary LLM with double‑calibration capability in a plug‑and‑play manner.

For a given question $\boldsymbol{Q}$, the proxy model generates a set of candidate evidence‑confidence pairs, i.e., $\hat{\mathcal{Z}}_{\boldsymbol{Q}}=\{ (\hat{z}_{\boldsymbol{Q}}^{(i)}, \hat{c}^{(i)}) \}_{i=1...K}$. Each $\hat{z}_{\boldsymbol{Q}}^{(i)}$ is grounded in $\mathcal{G}$, yielding factual reasoning paths that share the same confidence score $\hat{c}^{(i)}$. These paths and their confidences are verbalized into a natural-language context and integrated into prompts following prior verbalized UQ methods~\cite{DBLP:conf/emnlp/TianMZSRYFM23,DBLP:conf/iclr/XiongHLLFHH24}. Processing this enriched context, the LLM produces a final answer along with a well-calibrated prediction confidence.
This design establishes a traceable chain of confidence and achieves double calibration: the proxy first calibrates the external evidence confidence, which then informs and refines the LLM’s final prediction calibration through its own verbalized uncertainty estimation.

\section{Experiments}
\subsection{Experimental Settings}
\paragraph{Datasets and Evaluation Metrics.}
Following \cite{DBLP:conf/iclr/LuoLHP24,DBLP:conf/iclr/Li0025}, we evaluate our framework on two widely-adopted Knowledge Graph Question Answering (KGQA) benchmarks: \textbf{WebQSP}~\cite{DBLP:conf/acl/YihRMCS16} and \textbf{CWQ}~\cite{DBLP:conf/naacl/TalmorB18}. To measure prediction accuracy, we report \textbf{Hits}, \textbf{Recall}, and macro-averaged \textbf{F1} score. To assess the reliability of confidence estimates, we report the \textbf{Expected Calibration Error (ECE)}~\cite{DBLP:conf/icml/GuoPSW17} and the \textbf{Adaptive Calibration Error (ACE)}~\cite{nixon2019measuring}. ECE and ACE measure the expected absolute difference between empirical accuracy and predicted confidence using equal-width and adaptive equal-size bins, respectively.

\paragraph{Baselines.}
To rigorously evaluate our Double-Calibration mechanism, we construct \textbf{Single-Calibration (SingleCal)} baselines by extending reasoning paradigms with verbalized Uncertainty Quantification (UQ) methods, which elicit a self-reported confidence score alongside the predicted answer.
We select three state-of-the-art reasoning frameworks:
(i) The base \textbf{LLM Reasoner} without KG access;
(ii) \textbf{RoG}~\cite{DBLP:conf/iclr/LuoLHP24}, a KG-RAG method that grounds reasoning in retrieved relational paths;
(iii) \textbf{SubgraphRAG}~\cite{DBLP:conf/iclr/Li0025}, which retrieves and reasons over KG subgraphs.
Each framework is combined with three representative UQ prompting techniques: \textbf{Vanilla}~\cite{DBLP:conf/emnlp/TianMZSRYFM23}, \textbf{CoT}~\cite{DBLP:conf/nips/KojimaGRMI22}, and \textbf{Self-Probing}~\cite{DBLP:conf/iclr/XiongHLLFHH24}. {Prompt templates are detailed in Appendix~\ref{sec:appendix-uq-prompts}.}

\paragraph{Implementation Details.} 
All evaluated methods employ GPT-3.5-turbo~\cite{floridi2020gpt} as the primary reasoner unless otherwise specified, ensuring that performance differences are directly attributable to the calibration mechanism rather than the base LLM capability. The evidence proxy in our DoublyCal is implemented with Llama2-7B-Chat~\cite{DBLP:journals/corr/abs-2307-09288}, which is trained via the SFT then RL pipeline described in Sec.~\ref{sec:method-proxy}. {More details of experimental settings are provided in Appendix~\ref{sec:appendix-exp-details}.}

\subsection{Main Results}
\label{sec:main-exp}
Table \ref{tab:main_results} summarizes the comparative performance of our DoublyCal against all SingleCal baselines.

\paragraph{Superiority of Double-Calibration.}
DoublyCal achieves the best overall performance, consistently leading in all prediction metrics (Hits, Recall, F1) and the lowest ECE and ACE. Notably, it establishes a new standard for reliability, reducing the ECE and ACE to levels significantly lower than all SingleCal baselines. This result demonstrates that while KG-RAG methods can enhance LLM factuality, calibrating \textit{both} the external KG evidence and the final prediction is necessary for achieving reliable reasoning. Furthermore, the observed gains in prediction metrics indicate that evidence confidence helps improve the quality of the retrieved evidence and further unlocks the LLM's reasoning potential.

\paragraph{Calibrated Evidence as an Anchor for Verbalized UQ.}
The efficacy of verbalized UQ methods varies across reasoning backbones, with none proving universally dominant. This inconsistency arises because these methods may be subject to LLMs' inherent overconfidence. DoublyCal addresses this by supplying KG evidence with calibrated confidence, providing a reliable external anchor that refines the LLM’s uncertainty expression. Consequently, DoublyCal stabilizes all three UQ techniques and achieves the lowest ECE and ACE in nearly every configuration, showing that externally calibrated evidence improves confidence elicitation.

\paragraph{Controlled Enhancement via RL.}
The RL stage yields a significant performance gain, improving average F1 by $\sim$3.0 percentage points over the SFT-only. While prior work notes RL's risk of harming calibration~\cite{DBLP:journals/corr/abs-2509-04664}, our Bayesian confidence-aligned reward successfully mitigates this trade-off, resulting in only a minor and controlled variation in ECE and ACE. This confirms that our reward design effectively balances accuracy and calibration.

\begin{figure}[t]
\centering
  \includegraphics[width=.9\linewidth]{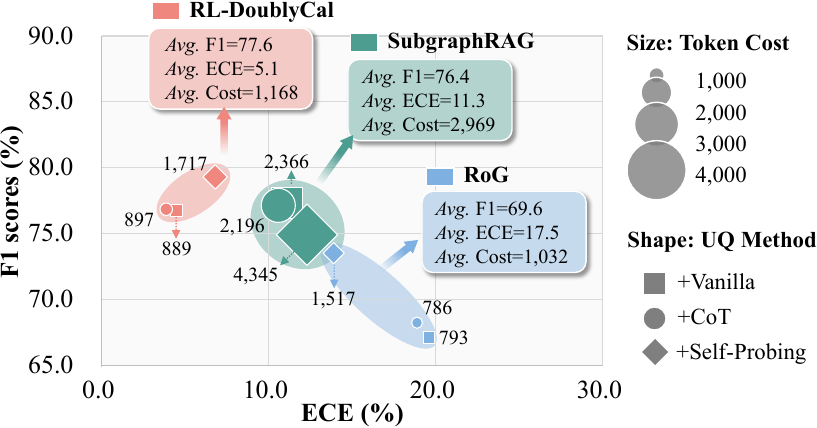}
  \caption{Input token efficiency analysis on WebQSP.}
  \label{fig:exp-efficiency}
\end{figure}

\subsection{Efficiency Analysis}
We further analyze the input token efficiency of each method, with results shown in Figure~\ref{fig:exp-efficiency}.

\paragraph{Superior Cost-Effectiveness of DoublyCal.}
DoublyCal substantially outperforms RoG (F1 +8.0, ECE -12.4) with only a marginal increase in input token cost, while consuming only about 39\% of the input tokens required by SubgraphRAG. This efficiency stems from the high information density of the evidence provided by DoublyCal. Specifically, compared to the simple relational paths in RoG, our constrained relational paths incorporate an optional constraint that yields more precise and informative evidence without compromising conciseness. Moreover, our proxy model is trained through evidence confidence calibration to select more discriminative KG evidence, further enhancing retrieval precision. In contrast, while SubgraphRAG’s subgraphs offer broader context, their lower information density leads to disproportionately high token costs.

\paragraph{Efficiency Across Different UQ Methods.}
Self-Probing incurs approximately twice the input token cost of Vanilla or CoT due to its two-step prompting design. However, because DoublyCal and RoG retrieve concise evidence, they can effectively leverage Self-Probing’s reflective ``second-thought'' process without excessive overhead. Notably, even when equipped with Self-Probing, their total input token cost remains below that of SubgraphRAG+Vanilla.

\begin{table}[t]
\centering
\resizebox{\linewidth}{!}{ % 
\begin{tabular}{cccccc}
\toprule[1.5pt]
\textbf{Variant} & \textbf{Hits} & \textbf{Recall} & \textbf{F1} & \textbf{ECE $\downarrow$} & \textbf{ACE $\downarrow$} \\
\midrule
\midrule
% --- SFT ---
\rowcolor{mygreen}\multicolumn{6}{c}{\textbf{SFT Only}} \\
\textbf{DoublyCal} & 90.0 & 81.0 & 72.6 & 3.1 & 3.8 \\
\textbf{SingleCal} & 90.1 \footnotesize{{(+0.1)}} & \cellcolor{mygray}80.4 \footnotesize{\textcolor{red}{(-0.6)}} & \cellcolor{mygray}72.5 \footnotesize{\textcolor{red}{(-0.1)}} & \cellcolor{mygray}21.2 \footnotesize{\textcolor{red}{(+18)}} & \cellcolor{mygray}20.3 \footnotesize{\textcolor{red}{(+16)}} \\
\textbf{Evidence} & \cellcolor{mygray}83.7 \footnotesize{\textcolor{red}{(-6.3)}} & \cellcolor{mygray}80.2 \footnotesize{\textcolor{red}{(-0.8)}} & \cellcolor{mygray}62.5 \footnotesize{\textcolor{red}{(-10)}} & \cellcolor{mygray}21.2 \footnotesize{\textcolor{red}{(+18)}} & \cellcolor{mygray}21.0 \footnotesize{\textcolor{red}{(+17)}} \\
\midrule
% --- RL ---
\rowcolor{mygreen}\multicolumn{6}{c}{\textbf{With RL}} \\
\textbf{DoublyCal} & 91.5 & 84.8 & 76.7 & 4.5 & 5.1 \\
\textbf{SingleCal} & 91.6 \footnotesize{{(+0.1)}} & \cellcolor{mygray}84.2 \footnotesize{\textcolor{red}{(-0.6)}} & \cellcolor{mygray}75.8 \footnotesize{\textcolor{red}{(-0.9)}} & \cellcolor{mygray}20.6 \footnotesize{\textcolor{red}{(+16)}} & \cellcolor{mygray}20.0 \footnotesize{\textcolor{red}{(+15)}} \\
\textbf{Evidence} & \cellcolor{mygray}86.4 \footnotesize{\textcolor{red}{(-5.1)}} & 84.8 & \cellcolor{mygray}67.8 \footnotesize{\textcolor{red}{(-8.9)}} & \cellcolor{mygray}21.3 \footnotesize{\textcolor{red}{(+16)}} & \cellcolor{mygray}21.1 \footnotesize{\textcolor{red}{(+16)}} \\
\bottomrule[1.5pt]
\end{tabular}
}
\caption{Ablation study results (\%) on the WebQSP dataset with the Vanilla UQ method. \colorbox{mygray}{Gray} marks performance degradation relative to the full model.}
\label{tab:exp-ablation}
\end{table}

\subsection{Ablation Analysis}
\label{sec:ablation}
We ablate DoublyCal against two variants (Table~\ref{tab:exp-ablation}) to examine the contribution of each component: (i) \textbf{SingleCal}, which removes the calibrated evidence confidence and applies calibration only to the final LLM output; (ii) \textbf{Evidence}, which removes the LLM reasoner and directly outputs the terminal entity of the factual path ($[\![z_{\boldsymbol{Q}}]\!]$) as the answer, using the evidence confidence as the final confidence score.

\paragraph{Evidence Confidence is Crucial for Final Prediction Calibration.}
Ablating from DoublyCal to SingleCal reveals a stark outcome: while predictive accuracy remains stable (e.g., F1 changes within $\pm$1 point), the calibration error (ECE and ACE) increases drastically from $\sim$4 to $>$20. This indicates that externally calibrated evidence confidence is essential, as it provides the LLM with a reliable anchor for its self-assessment. Without this first-stage calibration, the black-box LLM cannot reliably judge its own certainty, even when it can identify correct answers using high-quality KG evidence.

\paragraph{The LLM Reasoner Enables Integrative Reasoning.}
The significant performance gap of the Evidence variant underscores a pivotal design insight: the evidence proxy and the LLM reasoner play distinct yet complementary roles. The proxy specializes in evaluating individual KG evidence, while the LLM reasoner excels at synthesizing an ensemble of such evidence to perform complex reasoning. Consequently, the final prediction confidence is not a simple pass-through of any single evidence confidence, but rather the result of the LLM's holistic reasoning over the entire set of calibrated evidence. This efficient proxy-reasoner synergy is essential to the framework's performance.

\begin{figure}[t]
\centering
  \includegraphics[width=.88\linewidth]{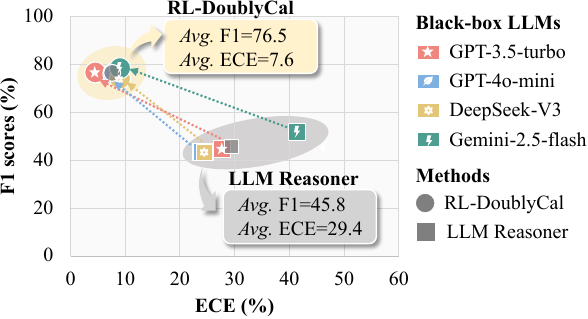}
  \caption{Prediction accuracy (F1) and calibration (ECE) of diverse black-box LLMs with and without DoublyCal.}
  \label{fig:exp-reasoner-all}
\end{figure}

\begin{figure}[t]
\centering
  \includegraphics[width=.92\linewidth]{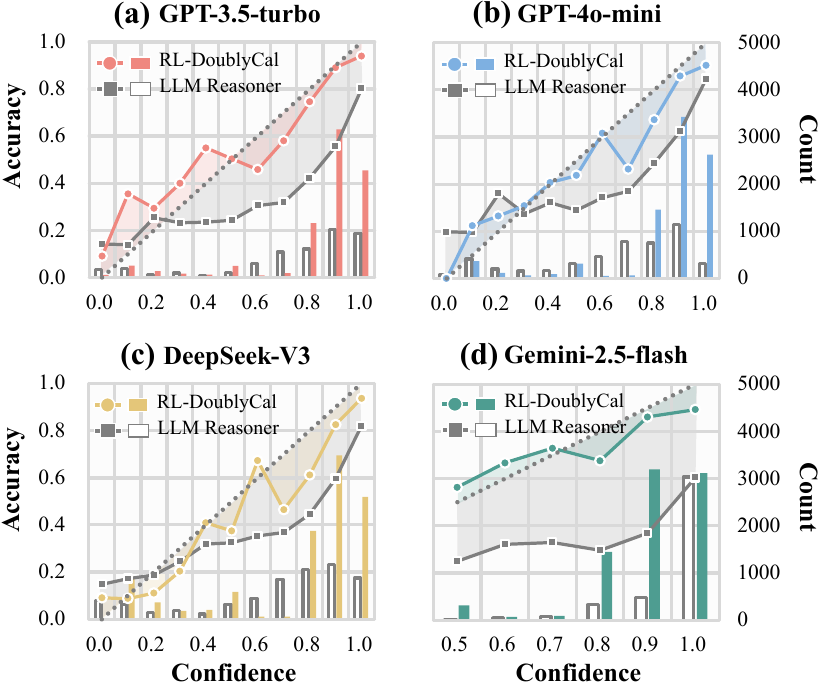}
  \caption{Calibration diagrams (\textit{bars}: confidence distribution per confidence bin; \textit{line}: empirical accuracy; \textit{dashed}: ideal calibration).}
  \label{fig:exp-reasoner}
\end{figure}

\begin{table}[t]
\centering
\resizebox{\linewidth}{!}{
\begin{tabular}{l}
\toprule[1.5pt]
\rowcolor{mygray}\multicolumn{1}{c}{\textbf{Sample}} \\
\textbf{Question:} Where did George W. Bush live as a child? \\
\textbf{Query Entity ($q$):} George W. Bush \quad\quad \textbf{Answers:}  New Haven. \\
\midrule
\midrule
\rowcolor{myorange}\multicolumn{1}{c}{\textbf{RL-DoublyCal} + Self-Probing} \\
\textbf{Retrieval:} \\
$\bullet$ $q$ $\rightarrow$ people.person.place\_of\_birth $\rightarrow$ New Haven [Confidence: 0.8] \\
$\bullet$ $q$ $\rightarrow$ people.person.place\_of\_birth $\rightarrow$ New Haven $\rightarrow$ \\
\quad location.location.containedby $\rightarrow$ \textit{Connecticut} [Confidence: 0.5] \\
$\bullet$ $q$ $\rightarrow$ people.person.place\_of\_birth $\rightarrow$ New Haven $\rightarrow$ \\
\quad location.location.containedby $\rightarrow$ United States of America [Confidence: 0.5] \\
\textbf{Predictions:}  \{\textit{Connecticut}: 0.3\} \\
% \midrule
% \midrule
% \rowcolor{mygreen}\multicolumn{1}{c}{\textbf{RoG} + Self-Probing} \\
% \textbf{Retrieval:} \\
% $\bullet$ $q$ $\rightarrow$ people.person.place\_of\_birth $\rightarrow$ New Haven \\
% $\bullet$ $q$ $\rightarrow$ people.place\_lived.person $\rightarrow$ m.03prwzr $\rightarrow$ \\
% \quad people.place\_lived.location $\rightarrow$ \textit{Midland} \\
% $\bullet$ $q$ $\rightarrow$ people.person.nationality $\rightarrow$ United States of America \\
% \quad $\rightarrow$ location.location.containedby $\rightarrow$ St. Louis \\
% $\bullet$ ... \\
% \textbf{Predictions:} \{\textit{Midland}: 0.9\} \\
\midrule
\midrule
\rowcolor{mygreen}\multicolumn{1}{c}{\textbf{SubgraphRAG} + Vanilla} \\
\textbf{Retrieval:} \\
$\bullet$ ($q$, people.person.place\_of\_birth, New Haven) \\
$\bullet$ ($q$, people.person.nationality, United States of America) \\
$\bullet$ (m.03prwzr, people.place\_lived.location, \textit{Midland}) \\ 
$\bullet$ (m.02xlp0j, people.place\_lived.location, Washington, D.C.) \quad $\bullet$ ... \\
% $\bullet$ ... \\
\textbf{Predictions:}  \{\textit{Midland}: 1.0\} \\
\bottomrule[1.5pt]
\end{tabular}
}
\caption{Case study: DoublyCal vs. SingleCal baseline.}
\label{tab:case-baseline}
\end{table}

\begin{table}[t]
\centering
\resizebox{\linewidth}{!}{
\begin{tabular}{l}
\toprule[1.5pt]
\rowcolor{mygray}\multicolumn{1}{c}{\textbf{Sample}} \\
\textbf{Question:} Where was Martin Luther King, Jr. raised? \\
\textbf{Query Entity ($q$):} Martin Luther King, Jr. \quad\quad \textbf{Answers:}  Atlanta. \\
\midrule
\midrule
\rowcolor{myorange}\multicolumn{1}{c}{\textbf{RL-DoublyCal} (Full) + Vanilla} \\
\textbf{Retrieval:} \\
$\bullet$ $q$ people.person.place\_of\_birth $\rightarrow$ \textit{Atlanta} [Confidence: 0.8] \\
$\bullet$ $q$ $\rightarrow$ people.deceased\_person.place\_of\_death $\rightarrow$ \textit{Memphis} [Confidence: 0.8] \\
\textbf{Predictions:}  \{\textit{Atlanta}: 0.8, \textit{Memphis}: 0.1\} \\
\midrule
\midrule
\rowcolor{mygreen}\multicolumn{1}{c}{\textbf{RL-DoublyCal} (SingleCal) + Vanilla} \\
\textbf{Retrieval:} Same factual paths as above, without confidence scores \\
\textbf{Predictions:}  \{\textit{Atlanta}: 0.8, \textit{Memphis}: 0.2\} \\
\bottomrule[1.5pt]
\end{tabular}
}
\caption{Case study: DoublyCal vs. its SingleCal variant.}
\label{tab:case-ablation}
\end{table}

\subsection{Cross-model Compatibility Analysis}
\label{sec:robustness}
To assess generalizability of DoublyCal, we evaluate it across diverse black-box LLMs, including GPT-3.5-turbo, GPT-4o-mini~\cite{achiam2023gpt}, DeepSeek-V3~\cite{liu2024deepseek}, and Gemini-2.5-flash~\cite{comanici2025gemini}.

\paragraph{Performance-Reliability Trade-off in LLM Reasoners.}
Figure \ref{fig:exp-reasoner-all} reveals a clear trade-off between accuracy and calibration among standalone LLMs. While GPT-family models and DeepSeek achieve comparable accuracy with moderate calibration errors (F1: 43.3–44.6; ECE: 23.9–27.7), Gemini attains a notably higher F1 (51.8) at the cost of a significantly worse ECE (41.4). This pattern highlights a common pitfall where optimizing purely for accuracy often degrades reliability in standalone LLMs. Confidence distributions (Figure \ref{fig:exp-reasoner}) confirms that all models exhibit systematic overconfidence. This issue is most acute in Gemini, where roughly 80\% of predictions are made with maximal confidence (1.0), yet the accuracy within this high-confidence group is only about 0.6.

\paragraph{DoublyCal Systematically Decouples the Trade-off.}
DoublyCal delivers consistent and substantial improvements across all models, effectively decoupling this trade-off. As shown in Figure \ref{fig:exp-reasoner-all}, it raises the average F1 from 45.8 to 76.5 while reducing the average ECE from 29.4 to 7.6. Crucially, DoublyCal mitigates the overconfidence patterns observed in black-box LLMs (Figure \ref{fig:exp-reasoner}), shifting confidence distributions toward well-calibrated and high-accuracy regions. By grounding confidence in externally calibrated evidence, DoublyCal provides a generalizable solution that enhances both accuracy and reliability of diverse black-box LLMs.

\subsection{Case Studies}
\label{sec:exp-case}
This section presents two case studies that qualitatively illustrate how DoublyCal enhances reliability of LLM predictions.

As shown in Table~\ref{tab:case-baseline}, for a question lacking a direct KG relation (``lived as a child'') to provide accurate support, both methods retrieve the related birthplace fact. However, while SubgraphRAG retrieves scattered evidence leading to an overconfident error (confidence 1.0), DoublyCal presents concise paths with calibrated confidence scores. Its birthplace path receives high confidence (0.8), while less relevant expansions are scored lower (0.5). Consequently, the LLM correctly assigns low confidence (0.3) to the incorrect answer ``Connecticut''. Furthermore, the ablation study in Table~\ref{tab:case-ablation} demonstrates that explicitly providing evidence confidence makes the LLM’s predicted confidence more concentrated on the correct answer. Together, these cases show DoublyCal’s ability to mitigate overconfidence and improve calibration. {Appendix~\ref{sec:apx-complementary-exp} provides complementary experiments.}

\section{Conclusion}
This paper establishes the principle of double-calibration for constructing a calibrated reasoning chain from KG evidence retrieval to final LLM prediction. We implement this principle in DoublyCal, a reliable KG-RAG framework that integrates plug-and-play verbalized uncertainty quantification, thereby enhancing the traceability and reliability of diverse black-box LLMs. Our work offers a concrete step toward building more reliable and transparent LLM systems, contributing to the advancement of trustworthy AI.

\section*{Limitations}
The proposed Double-Calibration principle is task-agnostic and readily adaptable to KG-RAG, and its effectiveness is well supported by our experiments. However, the robustness of DoublyCal still leaves room for improvement. On the one hand, while the Beta-Bernoulli confidence estimator can partially alleviate the effect of incomplete KGs via Bayesian smoothing, the resolution of its estimated confidence may decline when the KG is extremely sparse. On the other hand, although constrained relational paths provide strong generality, the fixed form of KG evidence somewhat limits DoublyCal's performance on complex questions, such as those involving aggregation or comparison. Future work with richer evidence forms or paradigms that iteratively explore the KG and dynamically construct evidence (e.g., via agentic planning) could further increase flexibility and reduce the reasoning burden placed on the black-box LLM.

\bibliographystyle{named}
\bibliography{ijcai26}

\appendix

\section{The Necessity of Bayesian Smoothing}
\label{apx:theory-prior}
As described in Sec.~\ref{sec:evidence}, we employ a Bayesian Beta-Bernoulli model to estimate KG evidence confidence. This appendix provides the justification for this design, demonstrating why the adopted Bayesian approach is essential for achieving robust estimates.

\paragraph{The Small-Sample Challenge.}
The structural sparsity inherent in knowledge graphs ($|\mathcal{E}| \ll |\mathcal{V}|^2 \times |\mathcal{R}|$) means that a given evidence $z_{\boldsymbol{Q}}$ for a question $\boldsymbol{Q}$ often retrieves only a small set of candidate answers. Formally, the candidate set size $n = |[\![z_{\boldsymbol{Q}}]\!]|$ is low. For instance, evidence such as $z: \mathrm{SiblingOf}(q, \hat{a})$ typically yields $n \leq 5$ candidates. Consequently, it is frequent to encounter extreme sampling outcomes where the number of correct candidates, $s = |[\![z_{\boldsymbol{Q}}]\!] \cap \mathcal{A}|$, is either $0$ or $n$. 

In such extreme situations with small samples, the standard maximum-likelihood estimator (MLE), $\hat{p}_{MLE} = s/n$, collapses to an absolute and potentially misleading value:
\begin{equation*}  
\hat{p}_{MLE}=
             \begin{cases}
             0, & s=0 \\  
             1, & s = n.
             \end{cases}  
\end{equation*}

\begin{figure*}[t]
\centering
  \includegraphics[width=\textwidth]{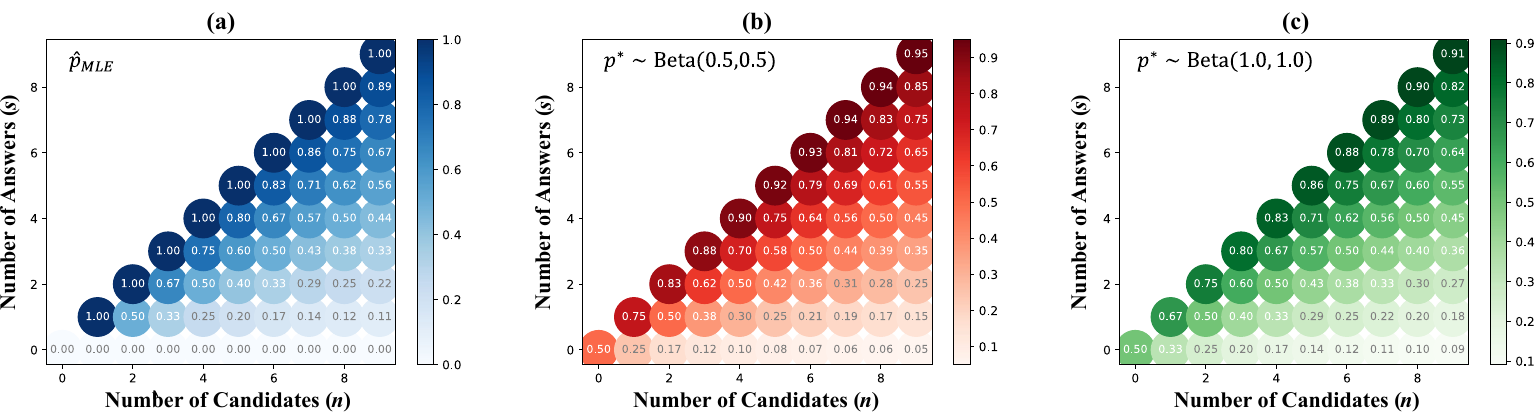}
  \caption{Comparison of \textbf{(a)} MLE, \textbf{(b)} Jeffreys-prior posterior mean, and \textbf{(c)} Uniform-prior posterior mean across varying candidate sizes $n$ and correct counts $s$.}
  \label{fig:beta-prior}
\end{figure*}

This behavior is illustrated in Figure~\ref{fig:beta-prior}(a). Such estimates are statistically unstable and semantically unreliable for real‑world KGs, often causing overconfidence in incorrect answers or undue dismissal of correct ones.

\paragraph{Bayesian Smoothing with the Jeffreys Prior.}
To mitigate this instability, we adopt a Bayesian approach. By introducing a conjugate Beta prior distribution for the parameter $p$, the point estimate is naturally shrunk toward the prior mean, thereby balancing the observed empirical frequency with a prior belief. A principled choice for this prior is the Jeffreys prior, $\mathrm{Beta}(0.5, 0.5)$~\cite{jeffreys1998theory}, which is theoretically well-motivated as a ``non-informative'' reference. Under this prior, the posterior mean estimator (Eq.~(4) in the main text) takes the form:
\begin{equation*}
p^*=\frac{0.5+s}{1+n}.
\end{equation*}

For the extreme cases, this becomes:
\begin{equation*}  
p^*=
             \begin{cases}
             \frac{0.5}{1+n}, & s=0, \\ 
             \addlinespace[5pt]
             \frac{0.5+n}{1+n}, & s = n.
             \end{cases}  
\end{equation*}

This formulation provides automatic and theoretically grounded smoothing. As shown in Figure~\ref{fig:beta-prior}(b), when $n$ is small, the estimate is conservatively pulled toward 0.5, guarding against overconfidence. As $n$ grows, $p^*$ converges to the MLE ($\hat{p}_{MLE}$), ensuring the estimator's consistency.

It is worth noting that the choice of prior controls the strength of this smoothing. For instance, the uniform prior $\mathrm{Beta}(1, 1)$~\cite{bayes1958essay} yields the estimator $p^* = (1 + s)/(2 + n)$ (Figure~\ref{fig:beta-prior}(c)), which exerts a stronger shrinkage effect toward $0.5$ than the Jeffreys prior and thus tends to produce more conservative estimates. These theoretical considerations are validated and their practical impact compared through an empirical analysis provided in Appendix~\ref{apx:exp-prior}.

\section{Prompt Templates for the Proxy Model}
\label{sec:appendix-prompt-temp}
This appendix details the input-output formats used for training the proxy model.

\subsection{Supervised Fine-Tuning (SFT)}
\label{sec:appendix-prompt-sft}
In the SFT stage, the proxy model learns to predict a target output sequence autoregressively. Each training instance consists of a natural-language instruction containing the question, paired with a target sequence that encodes the corresponding KG evidence in an XML-style format.

To formally describe the template, we define the symbolic placeholders: \texttt{[CONFIDENCE\_SCORE]} denotes the Bayesian confidence; \texttt{[RELATION\_PATH]} represents the core relational path, with individual relations separated by the special token \texttt{<SEP>}. Within the optional \texttt{<CONSTRAINT>} block, \texttt{[CONSTRAINED\_REL\_ENT]} signifies the concatenation of a constraining relation and its corresponding entity, also joined by \texttt{<SEP>}. The SFT template and a concrete example are provided below.

\subsection{Reinforcement Learning (RL)}
\label{sec:appendix-prompt-rl}
In the RL phase, the proxy model is prompted to generate an enhanced KG evidence path with well-calibrated confidence. The input instruction is adapted to encourage strategic decision-making, while the target output format remains identical to that used in the SFT stage.

\begin{tcolorbox}[title = {SFT Template}]
\textbf{Input:} \texttt{Please generate a valid relation path that can be helpful for answering the following question: [QUESTION]}

\tcbline

\textbf{Expected Output (with constraint):}

\texttt{<PATH confidence=[CONFIDENCE\_SCORE]> [RELATION\_PATH]<CONSTRAINT> [CONSTRAINED\_REL\_ENT] </CONSTRAINT></PATH>}

\tcbline

\textbf{Expected Output (without constraint):}

\texttt{<PATH confidence=[CONFIDENCE\_SCORE]> [RELATION\_PATH]</PATH>}

\end{tcolorbox}

\begin{tcolorbox}[title = {SFT Example}]
\textbf{Input:} 
\texttt{Please generate a valid relation path that can be helpful for answering the following question: what is the name of snoopy's brother?}

\tcbline

\textbf{Expected Output:} 

\texttt{<PATH confidence=0.75>sibling\_of <CONSTRAINT>gender<SEP>male </CONSTRAINT></PATH>}
\end{tcolorbox}

\begin{tcolorbox}[title = {RL Template}]
\textbf{Input:} \
\texttt{Please generate an enhanced relation path with well-calibrated confidence that can be helpful for answering the following question: [QUESTION]}
\end{tcolorbox}

\begin{tcolorbox}[title = {RL Example}]
\textbf{Input:} \
\texttt{Please generate an enhanced relation path with well-calibrated confidence that can be helpful for answering the following question: what is the name of snoopy's brother?}
\end{tcolorbox}

The proxy model must then generate a full output sequence (e.g., \texttt{<PATH confidence=...>...</PATH>}) based on this instruction. The generated sequence is subsequently evaluated by the reward function described in Sec.~\ref{sec:method-proxy}, which jointly assesses the inferential quality of the evidence path and the calibration accuracy of its attached confidence score.

\section{Training Details of the Proxy Model}
\label{sec:appendix-training-details}
This appendix details the training objectives and implementation for the SFT then RL training of the proxy model.

\subsection{SFT Stage}
\label{sec:appendix-sft-obj}
In the SFT stage, the proxy model $f_\theta$ is trained to autoregressively generate the target structured sequence (i.e., KG evidence with Bayesian confidence). The objective is to minimize the standard cross-entropy loss over the token sequence:
\begin{equation*}
\mathcal{L}_{\text{SFT}}(\theta) = - \sum\limits_{t=1}^{T} \log P_{\theta}(o_t \mid o_{<t}, \boldsymbol{Q}),
\end{equation*}
where $\boldsymbol{o}=(o_1, \dots, o_{T})$ is the token sequence of the target output, and $P_{\theta}(o_t \mid o_{<t}, \boldsymbol{Q})$ is the probability predicted by $f_{\theta}$ for the $t$-th token given the input question $\boldsymbol{Q}$ and previous tokens $o_{<t}$.

\subsection{RL Stage}
\label{sec:appendix-rl}
\paragraph{Final Reward Function.}
The reward defined in Eq.~(5) is a weighted sum of the inferential quality reward $R_{\text{inf}}$ and the calibration alignment reward $R_{\text{cal}}$, originally bounded in $[0,1]$. To stabilize optimization, we map the raw reward into the continuous interval $[-1,2]$ using a sigmoid-shaped transformation, which is defined as follows:
\begin{equation*}
\label{eq:smooth-reward}
R^\prime = 3 \cdot \sigma\bigl(\xi^\prime \cdot (R - 0.5)\bigr) - 1,
\end{equation*}
where $\sigma(\cdot)$ denotes the sigmoid function and $\xi^\prime > 0$ is a scaling hyperparameter (set to $\xi^\prime=2$ in our experiments). Additionally, we introduce a penalty of $-3$ for syntactically invalid outputs, such as when the generated sequence does not contain the required \texttt{<PATH>} tag.

\paragraph{GRPO Policy Objective.}
The proxy’s policy $\pi_{\theta}$ is optimized by minimizing the Group Relative Policy Optimization (GRPO) loss \cite{DBLP:journals/corr/abs-2402-03300}. This objective encourages higher reward while preventing excessive deviation from the reference policy (the model after SFT), thereby maintaining generation quality and training stability:
\begin{equation*}
\label{eq:grpo-loss}
\begin{aligned}
& \mathcal{L}_{\text{GRPO}}(\theta) =
-\frac{1}{G}
\sum_{i=1}^{G}
\mathbb{E}_{(s, \boldsymbol{o}_i)} \\
& \Bigg[
\frac{\pi_\theta(\boldsymbol{o}_i \mid s)}{\left[\pi_{\theta}(\boldsymbol{o}_i \mid s)\right]_{\text{no grad}}} \,
\hat{A}_i 
\;-\;
\beta^\prime \,
\text{KL}\big[ \pi_\theta \| \pi_{\text{ref}} \big]
\Bigg],
\end{aligned}
\end{equation*}
where $s$ denotes the shared input prompt for a group of size $G$, $\boldsymbol{o}_i$ is the $i$-th generated output sequence in the group, $\pi_{\text{ref}}$ is the reference policy (the model after SFT), $\hat{A}_i$ is the estimated advantage for the sequence $\boldsymbol{o}_i$, and $\beta^\prime$ controls the strength of the KL regularization term.

\section{Prompt Templates for UQ Methods}
\label{sec:appendix-uq-prompts}
This appendix details the prompt templates used for the three verbalized Uncertainty Quantification (UQ) methods evaluated in our work: \textbf{Vanilla}~\cite{DBLP:conf/emnlp/TianMZSRYFM23}, \textbf{CoT}~\cite{DBLP:conf/nips/KojimaGRMI22}, and \textbf{Self-Probing}~\cite{DBLP:conf/iclr/XiongHLLFHH24}. Each template is designed to elicit answers along with calibrated confidence estimates from a black-box LLM.

\subsection{Vanilla Template}
The Vanilla template directly instructs the model to output answers with confidence scores in a specified JSON format.

\begin{tcolorbox}[title = {Vanilla Template}]
\textbf{Input:} 
\texttt{[KG\_RAG\_INSTRUCTION] Please answer the following questions and \underline{provide the confidence} (0.0 to 1.0) for each answer being correct. Please keep the answer as simple as possible and return all the possible answers and their confidence as a json string.}

\texttt{Output format example: \{<answer\_1>: <confidence\_1>,...,<answer\_k>: <confidence\_k>\}}

\texttt{[KG\_RAG\_CONTEXT]}

\texttt{Question:
[QUESTION]}
\end{tcolorbox}

\subsection{Chain-of-Thought (CoT) Template}
The CoT template extends the Vanilla approach by appending the instruction ``\texttt{Let's think it step by step.}'' before presenting the context and question, thereby encouraging the model to generate an explicit reasoning chain prior to providing the final answer and confidence.

\begin{tcolorbox}[title = {CoT Template}]
\textbf{Input:} 
\texttt{[KG\_RAG\_INSTRUCTION] Please answer the following questions and provide the confidence (0.0 to 1.0) for each answer being correct. Please keep the answer as simple as possible and return all the possible answers and their confidence as a json string.}

\texttt{Output format example: \{<answer\_1>: <confidence\_1>,...,<answer\_k>: <confidence\_k>\}}

\texttt{\underline{Let's think it step by step.}}

\texttt{[KG\_RAG\_CONTEXT]}

\texttt{Question:
[QUESTION]}
\end{tcolorbox}

\subsection{Self-Probing Template}
The Self-Probing method employs a two-round dialogue. The first prompt elicits a list of candidate answers. The LLM's generated answer list is then used in a second prompt, which instructs it to analyze the likelihood of each answer being correct and to output the corresponding confidence scores in the same JSON format.

\begin{tcolorbox}[title = {Self-Probing Template}]
\textbf{First Interaction (Answer Generation):}
\texttt{[KG\_RAG\_INSTRUCTION] Please answer the given question. Please keep the answer as simple as possible and return all the possible answers as a list.}

\texttt{[KG\_RAG\_CONTEXT]}

\texttt{Question:
[QUESTION]}

\tcbline

\textbf{Model Output:} \texttt{[ANSWER\_LIST]}

\end{tcolorbox}

\begin{tcolorbox}[title = {Self-Probing Template (Cont.)}]

\textbf{Second Interaction (Confidence Elicitation):}
\texttt{Q: How likely are the above answers to be correct? Analyze the possible answers, provide your reasoning concisely, and give your confidence (0.0 to 1.0) for each answer being correct. Please keep the answer as simple as possible and return all the possible answers and their confidence as a json string.}

\texttt{Output format example: \{<answer\_1>: <confidence\_1>,...,<answer\_k>: <confidence\_k>\}}
\end{tcolorbox}

\section{Details of Experimental Settings}
\label{sec:appendix-exp-details}
\subsection{Datasets}
Our main experiments are conducted on two established KGQA benchmarks: \textbf{WebQSP}~\cite{DBLP:conf/acl/YihRMCS16} and \textbf{CWQ}~\cite{DBLP:conf/naacl/TalmorB18}, both of which are based on the Freebase knowledge graph~\cite{DBLP:conf/sigmod/BollackerEPST08}. To ensure fair comparison, we adopt the same train/validation/test splits used in prior work~\cite{DBLP:conf/iclr/LuoLHP24,DBLP:conf/iclr/Li0025}. The detailed statistics of both datasets are presented in Table~\ref{tab:exp-stat}.

\begin{table}[h]
\centering
\resizebox{\linewidth}{!}{
\begin{tabular}{@{}ccccc@{}}
\toprule[1.5pt]
\textbf{Dataset} & \textbf{\#Train} & \textbf{\#Validation} & \textbf{\#Test} & \textbf{Max \#Hop} \\
\midrule
\midrule
WebQSP & 2,826 & 225 & 1,628 & 2 \\
CWQ    & 27,639 & 2,577 & 3,531 & 4 \\
\bottomrule[1.5pt]
\end{tabular}
}
\caption{Statistics of the Freebase datasets.}
\label{tab:exp-stat}
\end{table}

To further verify the robustness of DoublyCal across different KG schemas, we evaluate it on the \textbf{MetaQA} dataset~\cite{zhang2018variational}, which is constructed from WikiMovies and employs a separate KG. The statistics of MetaQA are summarized in Table~\ref{tab:exp-metaqa-stat}. Due to computational resource constraints, we randomly sample 1,000 questions from the original test set of each MetaQA split for evaluation.

\begin{table}[h]
\centering
\resizebox{\linewidth}{!}{
\begin{tabular}{@{}ccccc@{}}
\toprule[1.5pt]
\textbf{Dataset} & \textbf{\#Train} & \textbf{\#Validation} & \textbf{\#Test} & \textbf{Sampled \#Test} \\
\midrule
\midrule
MetaQA-1hop & 96,106 & 9,992 & 9,947 & 1,000 \\
MetaQA-2hop & 118,980 & 14,872 & 14,872 & 1,000 \\
MetaQA-3hop & 114,196 & 14,274 & 14,274 & 1,000 \\
\bottomrule[1.5pt]
\end{tabular}
}
\caption{Statistics of the MetaQA datasets.}
\label{tab:exp-metaqa-stat}
\end{table}

\subsection{Evaluation Metrics}
To evaluate KGQA performance, we follow~\cite{DBLP:conf/iclr/LuoLHP24} and report \textbf{Hits}, \textbf{Recall}, and \textbf{F1}. For a question $\boldsymbol{Q}$ with gold answer set $\mathcal{A}$: 
\begin{itemize}
    \item \textbf{Hits} indicates whether the predicted answer set $\hat{\mathcal{A}}$ contains at least one correct answer:
    \begin{equation}
    \text{Hits} = \mathbf{1}\bigl[|\hat{\mathcal{A}} \cap \mathcal{A}| > 0\bigr],
    \end{equation}
    where $\mathbf{1}[\cdot]$ is the indicator function.
    \item \textbf{Recall} measures the fraction of gold answers covered by the prediction:
    \begin{equation}
    \text{Recall} = \frac{|\hat{\mathcal{A}} \cap \mathcal{A}|}{|\mathcal{A}|}.
    \end{equation}
    \item \textbf{F1} is the harmonic mean of Precision and Recall:
    \begin{equation}
    \text{F1} = 2 \cdot \frac{\text{Precision} \cdot \text{Recall}}{\text{Precision} + \text{Recall}},
    \end{equation}
    where $\text{Precision} = |\hat{\mathcal{A}} \cap \mathcal{A}| \,/\, |\hat{\mathcal{A}}|$ is the fraction of predicted answers that are correct.
\end{itemize}

To evaluate confidence calibration, we use the Expected Calibration Error (\textbf{ECE}) and Adaptive Calibration Error (\textbf{ACE}). Following the standard definition of \textbf{ECE} in~\cite{DBLP:conf/icml/GuoPSW17}, we partition all predictions into $M$ equal-width bins according to their predicted confidence. In our experiments, we set $M = 10$. Formally,
\begin{equation}
\text{ECE} = \sum_{m=1}^{M} \frac{|B_m|}{N} \bigl| \mathrm{acc}(B_m) - \mathrm{conf}(B_m) \bigr|,
\end{equation}
where $N$ is the total number of samples, $B_m$ denotes the set of predictions falling into the $m$-th bin, $\mathrm{acc}(B_m)$ denotes the accuracy within that bin, and $\mathrm{conf}(B_m)$ denotes the average predicted confidence in the bin.

\textbf{ACE} follows the same formulation but employs an adaptive binning scheme~\cite{nixon2019measuring}: predictions are sorted by confidence and partitioned into $M$ equal-size bins (we also use $M=10$).

\subsection{Implementation Details} 
\paragraph{SFT Training Dataset Construction.}
We construct the SFT training data through the following pipeline. For each question $\boldsymbol{Q}$, a breadth-first search (max depth$=4$) identifies the shortest relational paths $\mathcal{P}_r$ between the query entity $q$ and each candidate answer $a$. The Bayesian confidence for each $\mathcal{P}_r$ is computed as per Sec.~\ref{sec:evidence}. To enhance evidence quality, we then gather potential constraints $\mathcal{C}$ from each answer's one-hop neighborhood. For each candidate constrained path $\mathcal{P}_c = \mathcal{P}_r \wedge \mathcal{C}$, we compute its Bayesian confidence and retain it only if $\mathcal{P}_c$ yields a higher confidence than $\mathcal{P}_r$ alone. This filtering trains the proxy to identify genuinely valuable constraints. The SFT stage follows RoG's multi-task setup~\cite{DBLP:conf/iclr/LuoLHP24}, jointly optimizing the primary evidence generation and calibration task alongside an auxiliary QA task.

\paragraph{Training and Inference.}
The proxy model is trained using the AdamW optimizer~\cite{DBLP:conf/iclr/LoshchilovH19}. We set the learning rate to $2 \times 10^{-5}$ for the SFT stage and $1.41 \times 10^{-5}$ for the RL stage. Each stage is trained for at most 3 epochs with early stopping. All experiments were conducted on two NVIDIA A100 (80GB) GPUs. Training DoublyCal requires approximately 8.5 hours for SFT and 3 hours for RL, which is a lightweight one-time cost that enables efficient inference with any black-box LLM.

At inference time, the proxy model uses greedy decoding (temperature $\tau = 0$) to deterministically generate the top-$K$ evidence. Following RoG, we set $K = 3$, and the model produces constrained relational paths with a maximum depth of 4. For each question, this generation step takes around 1 second on an A100 GPU. The resulting paths are then grounded against the KG via efficient entity and relation lookups. This generative retrieval paradigm avoids expensive full‑graph traversal and scales sublinearly with KG size. 

\paragraph{Hyperparameter Selection.}
All key hyperparameters were set based on established design principles and empirical observations from pilot studies, given the computational cost of exhaustive grid search. For Bayesian calibration (Eq.~\eqref{eq:bayesian-map}), we adopt the weakly informative Jeffreys prior with $\alpha=\beta=0.5$~\cite{jeffreys1998theory}. In the RL reward function (Eq.~\eqref{eq:reward}), the balance weight $\lambda$ and tolerance $\xi$ are set to $0.85$ and $2$, respectively. For the GRPO objective, the KL regularization weight is $\beta' = 0.01$. This configuration proved stable and effective throughout all experiments.

\section{Complementary Experiments}
\label{sec:apx-complementary-exp}
\subsection{Significance Analysis}
\label{sec:apx-significance}
Bootstrap-based significance tests confirm that our RL-DoublyCal + Self-Probing significantly outperforms the strongest competitor (SubgraphRAG + Vanilla) on most metrics ($p<0.05$), whereas Recall on WebQSP is marginally significant, and only Hits on WebQSP and F1 on CWQ do not reach statistical significance. Crucially, DoublyCal achieves an order-of-magnitude reduction in ECE and ACE across all settings, validating its core strength in calibration. The accompanying accuracy gains are a welcome byproduct of improved reliability rather than our primary goal. The ablation study (Table~\ref{tab:exp-ablation}) clarifies the underlying mechanism: compared with its SingleCal variant that shares the same proxy and evidence but omits evidence‑confidence verbalization, DoublyCal yields comparable F1 while reducing ECE from $\sim$20 to $\sim$4. This demonstrates that evidence confidence indirectly improves reasoning accuracy by encouraging the proxy to prioritize high‑confidence evidence, and directly enhances calibration by informing the black‑box LLM.

\subsection{Evaluation on the MetaQA Datasets}
\label{sec:apx-metaqa}
Table~\ref{tab:metaqa} reports the results on MetaQA for \textbf{DoublyCal}, its ablated variant \textbf{SingleCal} (which omits evidence confidence), and the \textbf{LLM Reasoner} without knowledge augmentation. All experiments use GPT-3.5-turbo as the black-box LLM with the Vanilla UQ method.

\paragraph{DoublyCal exhibits robustness across KG schemas.} 
Consistent with its competitive performance on WebQSP and CWQ, our RL-DoublyCal attains leading predictive performance across all three MetaQA splits while achieving substantial calibration improvements (e.g., F1 $>$ 85 and ECE $<$ 7).
This confirms that the theoretically grounded Bayesian confidence estimator, combined with the generative capacity of an effectively trained LM-based proxy model, constitutes a reliable reasoning solution that generalizes effectively across distinct KGs.

\paragraph{The advantage of DoublyCal becomes more pronounced on harder questions.} 
On 1-hop and 2-hop questions, SingleCal matches DoublyCal in predictive performance yet exhibits markedly inferior calibration.
Moreover, both the LLM Reasoner and SingleCal undergo sharp performance declines as question difficulty increases, whereas DoublyCal remains substantially more robust.
On the most challenging 3-hop split, RL-DoublyCal attains F1 $=$ 85.4 and ECE $=$ 5.9, surpassing the strongest RL-SingleCal baseline by 5.1 F1 points and 51.2 ECE points.
Notably, DoublyCal’s low calibration error on hard questions demonstrates its effectiveness in mitigating the overconfidence problem typical of LLMs.

\begin{table}[h]
\centering
\resizebox{\linewidth}{!}{%
\begin{tabular}{lccccc}
\toprule[1.5pt]
\textbf{Method} & \textbf{Hits} & \textbf{Recall} & \textbf{F1} & \textbf{ECE} $\downarrow$ & \textbf{ACE} $\downarrow$ \\
\midrule
\midrule
\rowcolor{mygray}\multicolumn{6}{c}{\textbf{MetaQA-1hop}} \\
\addlinespace[2pt]
LLM-Reasoner   & 62.6 & 54.8 & 37.4 & 34.4 & 34.8 \\
\addlinespace[2pt]
SFT-SingleCal  & 97.5 & 95.9 & 91.6 & 12.9 & 12.6 \\
SFT-DoublyCal  & \cellcolor{lightblue!25}97.7 & \cellcolor{lightblue!25}96.0 & \cellcolor{lightblue!25}92.1 & \cellcolor{lightblue!25}4.0 & \cellcolor{lightblue!25}3.7 \\
\addlinespace[2pt]
RL-SingleCal   & 99.3 & 98.9 & 95.3 & 16.4 & 16.9 \\
RL-DoublyCal   & 98.4 & 98.1 & 95.1 & \cellcolor{lightblue!25}6.7 & \cellcolor{lightblue!25}5.9 \\
\midrule
\rowcolor{mygray}\multicolumn{6}{c}{\textbf{MetaQA-2hop}} \\
\addlinespace[2pt]
LLM-Reasoner   & 34.9 & 22.1 & 15.7 & 43.4 & 43.2 \\
\addlinespace[2pt]
SFT-SingleCal  & 97.9 & 96.6 & 94.7 & 39.8 & 41.8 \\
SFT-DoublyCal  & 97.6 & \cellcolor{lightblue!25}96.7 & \cellcolor{lightblue!25}96.1 & \cellcolor{lightblue!25}3.9 & \cellcolor{lightblue!25}3.7 \\
\addlinespace[2pt]
RL-SingleCal   & 99.5 & 98.5 & 97.7 & 43.7 & 44.1 \\
RL-DoublyCal   & 99.3 & 98.1 & \cellcolor{lightblue!25}97.9 & \cellcolor{lightblue!25}1.2 & \cellcolor{lightblue!25}1.2 \\
\midrule
\rowcolor{mygray}\multicolumn{6}{c}{\textbf{MetaQA-3hop}} \\
\addlinespace[2pt]
LLM-Reasoner   & 51.6 & 19.8 & 19.3 & 28.3 & 28.3 \\
\addlinespace[2pt]
SFT-SingleCal  & 94.5 & 80.6 & 74.2 & 50.4 & 54.2 \\
SFT-DoublyCal  & \cellcolor{lightblue!25}94.8 & \cellcolor{lightblue!25}84.7 & \cellcolor{lightblue!25}78.8 & \cellcolor{lightblue!25}4.9 & \cellcolor{lightblue!25}3.9 \\
\addlinespace[2pt]
RL-SingleCal   & 95.6 & 85.0 & 80.3 & 57.2 & 61.7 \\
RL-DoublyCal   & \cellcolor{lightblue!25}96.9 & \cellcolor{lightblue!25}90.4 & \cellcolor{lightblue!25}85.4 & \cellcolor{lightblue!25}5.9 & \cellcolor{lightblue!25}5.7 \\
\bottomrule[1.5pt]
\end{tabular}
}
\caption{Results (\%) on the MetaQA dataset. \colorbox{lightblue!25}{Light‑blue cells} in a DoublyCal row indicate improvement over the corresponding SingleCal variant.}
\label{tab:metaqa}
\end{table}

\subsection{Sensitivity Analysis of the Proxy Base Model}
We examine the sensitivity of DoublyCal to the scale of the proxy model by replacing the default Llama2-7B-Chat backbone with two substantially smaller language models: Qwen3‑4B and Qwen3‑1.7B~\cite{qwen3technicalreport}.
All proxy models are trained with SFT followed by RL, and evaluated with the Vanilla UQ method on WebQSP.
As reported in Table~\ref{tab:proxy-base}, Llama2‑7B exhibits a marginal advantage in predictive metrics (e.g., +1.3 Hits over Qwen3‑4B), yet the overall performance of the three proxy backbones remains closely matched across both accuracy and calibration dimensions.
These results demonstrate that DoublyCal is robust to significant reductions in proxy model capacity, and it can deliver strong KGQA accuracy together with reliable confidence estimates even under tight computational constraints.

\begin{table}[h]
\centering
\begin{tabular}{@{}lccccc@{}}
\toprule[1.5pt]
\textbf{Base Model} & \textbf{Hits} & \textbf{Recall} & \textbf{F1} & \textbf{ECE} $\downarrow$ & \textbf{ACE} $\downarrow$ \\
\midrule
\midrule
Llama2-7B  & 91.5 & 84.8 & 76.7 & 4.6 & 5.1 \\
Qwen3-4B   & 90.2 & 83.7 & 75.3 & 5.7 & 5.5 \\
Qwen3-1.7B & 90.2 & 84.2 & 76.3 & 4.6 & 4.8 \\
\bottomrule[1.5pt]
\end{tabular}
\caption{Performance of DoublyCal + Vanilla with different proxy base models on WebQSP.}
\label{tab:proxy-base}
\end{table}

\subsection{Empirical Analysis of Prior Selection}
\label{apx:exp-prior}
Building upon the theoretical justification in Appendix~\ref{apx:theory-prior}, we provide an empirical comparison of different prior choices for Bayesian evidence estimation and analyze their impact on the overall performance of our DoublyCal framework.

The results in Table~\ref{tab:exp-prior} demonstrate the trade-off between accuracy and calibration controlled by the prior. Empirically, the use of MLE (equivalent to the Haldane prior $\mathrm{Beta}(0,0)$~\cite{haldane1932note}) leads to severe calibration degradation, yielding the highest F1 score (75.3) but also the worst ECE (13.4). This aligns with the theoretical risk that point estimates in small-sample extremes produce overconfident outputs, undermining system reliability. While the Uniform prior $\mathrm{Beta}(1, 1)$~\cite{bayes1958essay} applies stronger shrinkage, the Jeffreys prior $\mathrm{Beta}(0.5, 0.5)$~\cite{jeffreys1998theory} provides a better balance, achieving by far the best calibration (ECE 3.1) while maintaining a competitive F1 score (72.6).

\begin{table}[h]
\centering
\resizebox{\linewidth}{!}{ % 调整了缩放比例以适应新列宽
\begin{tabular}{ccccccc}
\toprule[1.5pt]
\textbf{Prior Variant} & $\alpha$ & $\beta$ & \textbf{Hits} & \textbf{Recall} & \textbf{F1} & \textbf{ECE $\downarrow$} \\
\midrule
\midrule
\textbf{MLE} & 0.0 & 0.0 & 91.0 & 81.4 & 75.3  & 13.4 \\
\textbf{Jeffreys prior} & 0.5 & 0.5 & 90.0 & 81.0 & 72.6 & 3.1 \\
\textbf{Uniform prior} & 1.0 & 1.0 & 89.6 & 81.8 & 74.0 & 6.8 \\
\bottomrule[1.5pt]
\end{tabular}
}
\caption{Performance (\%) of SFT-DoublyCal+Vanilla with different priors for Bayesian estimation on WebQSP.}
\label{tab:exp-prior}
\end{table}

\subsection{Impact of UQ on Predictive Accuracy}
\label{sec:exp-acc-vs-reliability}
To complement the main experiments, we investigate whether prompting the LLM to verbalize its uncertainty affects predictive accuracy in KGQA. We compare recent KGQA methods with our strongest baselines and DoublyCal without explicit uncertainty quantification (UQ).

Table~\ref{tab:exp-uq-tradeoff} reveals a contrasting pattern. For standard KG-RAG baselines (RoG and SubgraphRAG), verbalized UQ consistently improves F1 scores. This supports the view that uncertainty prompts can induce more deliberate reasoning, which is beneficial for complex questions.
In contrast, our RL-DoublyCal already achieves top-tier accuracy without any UQ prompt. Notably, adding UQ with RL-DoublyCal brings no substantial gain and can even slightly lower F1. A plausible explanation is that the high quality of the KG evidence selected by DoublyCal’s proxy lowers the intrinsic reasoning difficulty for the primary LLM, thereby reducing the marginal benefit of an extra ``second thought'' prompted by UQ. This inherent strength positions DoublyCal to better balance the dual objectives of high accuracy and reliable calibration when UQ is employed, as demonstrated by its consistently low ECE in the main results (Table~\ref{tab:main_results}).

\begin{table}[h]
\centering
\resizebox{\linewidth}{!}{
\begin{tabular}{@{}lcc@{}}
\toprule[1.5pt]
\textbf{Reasoning Method} & \textbf{WebQSP} & \textbf{CWQ} \\
\midrule
\midrule
\textbf{SR+NSM}~\cite{DBLP:conf/acl/ZhangZY000C22} & 64.1 & 47.1 \\
\textbf{SR+NSM+E2E}~\cite{DBLP:conf/acl/ZhangZY000C22} & 64.1 & 46.3 \\
\textbf{UniKGQA}~\cite{DBLP:conf/iclr/JiangZ0W23} & 72.2 & 49.0 \\
\textbf{G-Retriever}~\cite{DBLP:conf/nips/He0SC0LBH24} & 73.5 & - \\
\textbf{GNN-RAG}~\cite{DBLP:conf/acl/MavromatisK25} & 71.3 & 59.4 \\
\midrule
\textbf{RoG} (GPT-3.5-turbo) & 66.8 & 	46.5 \\
\quad +UQ (Self-Probing) & 73.5 & 48.7 \\
\midrule
\textbf{SubgraphRAG} (GPT-3.5-turbo) &	74.7  &	52.1  \\
\quad +UQ (Vanilla) & 77.3 & 52.2 \\
\midrule
\midrule
\textbf{RL-DoublyCal} (GPT-3.5-turbo) &	79.7 & 	52.1  \\
\quad +UQ (Self-Probing) & 	79.3 &	53.0 \\
\bottomrule[1.5pt]
\end{tabular}
}
\caption{F1 scores (\%) of reasoning methods without and with UQ.}
\label{tab:exp-uq-tradeoff}
\end{table}

\begin{table*}[t]
\centering
\resizebox{\linewidth}{!}{
\begin{tabular}{ll}
\toprule[1.5pt]
\rowcolor{mygray}\multicolumn{2}{c}{\textbf{Sample}} \\
\textbf{Question:} & Where did George W. Bush live as a child? \\
\textbf{Answers:} &  New Haven. \\
\midrule
\midrule
\rowcolor{myorange}\multicolumn{2}{c}{\textbf{RL-DoublyCal} + Self-Probing} \\
\textbf{Retrieval:} &  George W. Bush $\rightarrow$ people.person.place\_of\_birth $\rightarrow$ New Haven [Confidence: 0.8] \\
& George W. Bush $\rightarrow$ people.person.place\_of\_birth $\rightarrow$ New Haven $\rightarrow$ location.location.containedby $\rightarrow$ Connecticut [Confidence: 0.5] \\
& George W. Bush $\rightarrow$ people.person.place\_of\_birth $\rightarrow$ New Haven $\rightarrow$ location.location.containedby $\rightarrow$ United States of America [Confidence: 0.5] \\
\textbf{Predictions:} &  \{Connecticut: 0.3\} \\
\midrule
\midrule
\rowcolor{mygreen}\multicolumn{2}{c}{\textbf{RoG} + Self-Probing} \\
\textbf{Retrieval:} & George W. Bush $\rightarrow$ people.person.place\_of\_birth $\rightarrow$ New Haven \\
&  George W. Bush $\rightarrow$ people.place\_lived.person $\rightarrow$ m.03prwzr $\rightarrow$ people.place\_lived.location $\rightarrow$ Midland \\
& George W. Bush $\rightarrow$ people.person.nationality $\rightarrow$ United States of America $\rightarrow$ location.location.containedby $\rightarrow$ St. Louis ... \\
\textbf{Predictions:} & \{Midland: 0.9\} \\
\midrule
\midrule
\rowcolor{mygreen}\multicolumn{2}{c}{\textbf{SubgraphRAG} + Vanilla} \\
\textbf{Retrieval:} & (George W. Bush, people.person.place\_of\_birth, New Haven)\quad (George W. Bush, people.person.nationality, United States of America) ... \\
& (m.03prwzr, people.place\_lived.location, Midland)\quad (m.02xlp0j, people.place\_lived.location, Washington, D.C.) ... \\
\textbf{Predictions:} &  \{Midland: 1.0\} \\
\bottomrule[1.5pt]
\end{tabular}
}
\caption{Comparative case study: DoublyCal vs. baselines on calibration.}
\label{tab:apx-case-baseline}
\end{table*}

\begin{table*}[t]
\centering
\resizebox{\linewidth}{!}{
\begin{tabular}{ll}
\toprule[1.5pt]
\rowcolor{mygray}\multicolumn{2}{c}{\textbf{Sample}} \\
\textbf{Question:} & Where was Martin Luther King, Jr. raised? \\
\textbf{Answers:} &  Atlanta. \\
\midrule
\midrule
\rowcolor{myorange}\multicolumn{2}{c}{\textbf{RL-DoublyCal} (Full) + Vanilla} \\
\textbf{Retrieval:} & Martin Luther King, Jr. $\rightarrow$ people.person.place\_of\_birth $\rightarrow$ Atlanta [Confidence: 0.8] \\
& Martin Luther King, Jr. $\rightarrow$ people.deceased\_person.place\_of\_death $\rightarrow$ Memphis [Confidence: 0.8] \\
\textbf{Predictions:} & \{Atlanta: 0.8, Memphis: 0.1\} \\
\midrule
\midrule
\rowcolor{mygreen}\multicolumn{2}{c}{\textbf{RL-DoublyCal} (SingleCal) + Vanilla} \\
\textbf{Retrieval:} & Martin Luther King, Jr. $\rightarrow$ people.person.place\_of\_birth $\rightarrow$ Atlanta \\
& Martin Luther King, Jr. $\rightarrow$ people.deceased\_person.place\_of\_death $\rightarrow$ Memphis \\
\textbf{Predictions:} &  \{Atlanta: 0.8, Memphis: 0.2\} \\
\midrule
\midrule
\\
\midrule
\midrule
\rowcolor{myorange}\multicolumn{2}{c}{\textbf{SFT-DoublyCal} (Full) + Vanilla} \\
\textbf{Retrieval:} & Martin Luther King, Jr. $\rightarrow$ people.person.place\_of\_birth $\rightarrow$ Atlanta [Confidence: 0.8] \\
& Martin Luther King, Jr. $\rightarrow$ people.person.nationality $\rightarrow$ United States of America $\rightarrow$ location.country.capital $\rightarrow$ Washington, D.C. [Confidence: 0.8] \\
\textbf{Predictions:} &  \{Atlanta: 0.8, United States of America: 0.2\} \\
\midrule
\midrule
\rowcolor{mygreen}\multicolumn{2}{c}{\textbf{SFT-DoublyCal} (SingleCal) + Vanilla} \\
\textbf{Retrieval:} & Martin Luther King, Jr. $\rightarrow$ people.person.place\_of\_birth $\rightarrow$ Atlanta \\
& Martin Luther King, Jr. $\rightarrow$ people.person.nationality $\rightarrow$ United States of America $\rightarrow$ location.country.capital $\rightarrow$ Washington, D.C.] \\
\textbf{Predictions:} &  \{Atlanta: 0.7, United States of America: 0.3\} \\
\bottomrule[1.5pt]
\end{tabular}
}
\caption{Ablation case study: Full vs. SingleCal variant.}
\label{tab:apx-case-ablation}
\end{table*}

\subsection{Case Studies}
This appendix provides an extended analysis of the cases presented in Sec.~\ref{sec:exp-case}.

\paragraph{Comparison with Baselines.}
Table~\ref{tab:apx-case-baseline} compares RL‑DoublyCal with the strongest baselines on the question ``Where did George W. Bush live as a child?''. None of the methods retrieves an exact supporting fact, because the KG lacks the explicit relation ``lived as a child''. All methods do retrieve the related fact ``George W. Bush was born in New Haven''. However, whereas DoublyCal presents concise factual paths accompanied by calibrated confidence scores, the evidence retrieved by RoG and SubgraphRAG is more scattered. Consequently, the LLMs guided by RoG and SubgraphRAG are distracted from the core entities (``George W. Bush'' and ``New Haven'') and assign overconfident scores (0.9–1.0) to the incorrect prediction ``Midland''.

In contrast, DoublyCal attaches calibrated confidence scores through its first-stage calibration. Specifically, the precise path about birthplace receives a high score (0.8), while the more generic expansions receive lower scores (0.5), reflecting their weaker inferential relevance. Guided by these scores, the black-box LLM correctly assigns low confidence (0.3) to the plausible but incorrect prediction ``Connecticut'', demonstrating better-calibrated uncertainty estimation.

\paragraph{Comparison with Ablated Models.}
Table~\ref{tab:apx-case-ablation} contrasts the full DoublyCal framework with its SingleCal ablation, which removes the calibrated evidence confidence (i.e., only the second-stage calibration remains). Both RL‑ and SFT‑DoublyCal retrieve high‑confidence evidence focused on the question (e.g., ``Martin Luther King, Jr. was born in Atlanta''). RL‑DoublyCal exhibits slightly sharper calibration, likely because its reward‑driven training promotes more discriminative evidence selection. More importantly, when evidence confidence is provided (the full model), the LLM’s predicted confidence is more concentrated on the correct answer. For example, RL-DoublyCal (Full) assigns only 0.1 confidence to the distracting alternative ``Memphis'', whereas its SingleCal variant assigns 0.2. Similarly, SFT-DoublyCal (Full) assigns 0.2 to ``United States of America'', while the SingleCal variant assigns 0.3. This directly demonstrates that the first-stage evidence calibration is crucial for providing a reliable confidence anchor, enabling the LLM to synthesize multiple evidence pieces into a decisive and well‑calibrated final prediction.

\end{document}